\documentclass{article}

\usepackage{iclr2027_conference,times}
\usepackage{amsmath,amssymb,booktabs,multirow,float,graphicx,microtype,pgfplots,pgf-pie,tikz,xcolor}
\usetikzlibrary{arrows.meta,patterns,positioning}
\usepgfplotslibrary{fillbetween,groupplots}
\pgfplotsset{compat=1.18}
\usepackage{subcaption,needspace}
\usepackage{flafter}
\usepackage{placeins}
\usepackage[colorlinks=true,linkcolor=blue,citecolor=blue,urlcolor=blue]{hyperref}

\usepackage{url}

\title{When Does Longer Reasoning Help? Predicting Mathematical Reasoning Through Discovery and Execution}
\author{ Adib Hasan \\
Independent Researcher \\
notadib@alum.mit.edu \And Lay Jain \\
Independent Researcher \\
layjain@alum.mit.edu
\And Thanic Nur Samin \\
IU Bloomington \\
tsamin@iu.edu
}

\newif\ifpreprint
\preprinttrue 
\ifpreprint
  \iclrfinalcopy
\fi

\begin{document}
\maketitle
\ifpreprint
  \lhead{Preprint}
\fi

\begin{abstract}

Test-time compute can improve mathematical reasoning, but can short-budget runs predict how mathematical reasoning scales with additional compute? We introduce a \textit{Discovery--Execution (DE)} framework that predicts the aggregate held-out scaling curves through a convolution of strategy \emph{discovery} and conditional \emph{execution}. From independent short-budget attempts and oracle-sketch-conditioned runs, the framework estimates cumulative success along held-out reasoning trajectories under alternate compute allocations. We evaluate four models on 35 fresh Olympiad problems and non-geometry problems from IMO-ProofBench Advanced. Under the DE framework, near-saturated execution predicts geometric scaling, as observed for the GPT models. For Claude Opus 4.8, incorporating measured execution substantially improves held-out forecasts over geometric extrapolation across one- and two-arm allocations. As a secondary application, regularized DE (R-DE) decisions to continue or restart yield lower average regret than the best model-specific retrospective policy. Together, these results show that measuring conditional execution provides information about longer reasoning that short-budget success rates do not always capture.
\end{abstract}

\section{Introduction}

\begin{figure}[t]
\centering
\begingroup
\definecolor{fwblue}{HTML}{356AA0}
\definecolor{fworange}{HTML}{D97706}
\definecolor{fwpurple}{HTML}{7A5195}
\definecolor{fwink}{HTML}{263442}
\begin{tikzpicture}[ x=1cm,y=1cm,font=\sffamily\small,text=fwink, flow/.style={-{Stealth[length=2mm,width=1.4mm]},line width=1pt}, tile/.style={rounded corners=0.6pt,line width=0.7pt},
]
\node[font=\sffamily\small\bfseries] at (1.85,2.05) {Measure};
\node[font=\sffamily\small\bfseries] at (6.55,2.05) {Framework};
\node[font=\sffamily\small\bfseries] at (11.3,2.05) {Predict};

\node[anchor=west,text=fwblue] at (0.2,1.58) {$N=8\quad K=1$};
\foreach \row in {0,1}{
  \foreach \col in {0,...,3}{
    \draw[tile,draw=fwblue,fill=fwblue!12] (0.25+0.68*\col,0.84-0.65*\row) rectangle ++(0.4,0.4);
  }
}
\node[text=fwblue,font=\footnotesize] at (3.12,0.72) {$1\times$};

\node[anchor=west,text=fworange] at (0.2,-0.57) {Oracle sketch};
\begin{scope}[shift={(0.25,-1.52)}]
  \draw[tile,draw=fworange,fill=fworange!7] (0,0) rectangle (0.5,0.64);
  \foreach \y in {0.16,0.32,0.48}{
    \fill[fworange] (0.1,\y) circle (0.025);
    \draw[fworange,line width=0.65pt] (0.18,\y)--(0.4,\y);
  }
\end{scope}
\draw[flow,fworange] (0.88,-1.2)--(1.19,-1.2);
\foreach \i in {0,...,7}{
  \draw[tile,draw=fworange,fill=fworange!12] (1.3+0.28*\i,-1.37) rectangle ++(0.28,0.34);
}

\draw[flow,fwblue] (3.8,0.72)--(4.65,0.72)--(5.215,0.18);
\node[text=fwblue,above] at (4.52,0.76) {$\widehat p(1)\,\widehat\varepsilon(1)$};
\draw[flow,fworange] (3.8,-1.2)--(4.65,-1.2)--(5.215,-0.66);
\node[text=fworange,below] at (4.52,-1.26) {$\widehat\varepsilon(\ell)$};
\draw[draw=fwpurple!50,fill=fwpurple!5,line width=0.8pt] (6.55,-0.24) circle (1.4);
\node[text=fwpurple] at (6.55,0.2) {$s=p\ast\varepsilon$};
\node[text=fwpurple] at (6.55,-0.4) {$1-(1-s(K))^N$};

\draw[draw=fwpurple,line width=1pt] (7.95,-0.24)--(8.35,-0.24);
\draw[flow,draw=fwpurple] (8.35,-0.24)|-(9.43,0.95);
\draw[flow,draw=fwpurple] (8.35,-0.24)|-(9.43,-0.25);
\draw[flow,draw=fwpurple] (8.35,-0.24)|-(9.43,-1.45);

\node[anchor=west] at (9.65,1.5) {$N=1\quad K=8$};
\foreach \i in {0,...,7}{
  \draw[tile,draw=fwpurple,fill=fwpurple!12] (9.65+0.34*\i,0.76) rectangle ++(0.34,0.38);
}
\node[anchor=west] at (9.65,0.3) {$N=2\quad K=4$};
\foreach \cluster in {0,1}{
  \foreach \i in {0,...,3}{
    \draw[tile,draw=fwpurple,fill=fwpurple!12] (9.65+1.52*\cluster+0.34*\i,-0.44) rectangle ++(0.34,0.38);
  }
}
\node[anchor=west] at (9.65,-0.9) {$N=4\quad K=2$};
\foreach \cluster in {0,...,3}{
  \foreach \i in {0,1}{
    \draw[tile,draw=fwpurple,fill=fwpurple!12] (9.65+0.84*\cluster+0.34*\i,-1.64) rectangle ++(0.34,0.38);
  }
}
\end{tikzpicture}
\endgroup
\caption{\textbf{Predicting the value of longer reasoning.} Short budget runs measure success within one block; runs supplied with an oracle strategy sketch measure execution. Together, these measurements estimate discovery and execution, whose convolution predicts held-out unaided success across compute allocations. Each tile is one block (200k output tokens); $N$ independent attempts receive $K$ blocks each, with $NK=8$.}
\label{fig:framework-overview}
\end{figure}
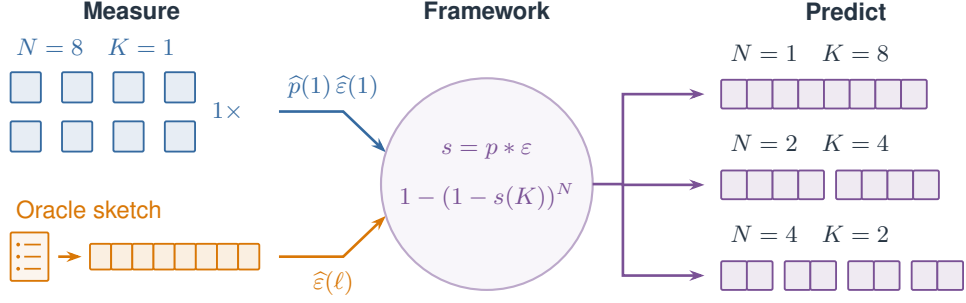

Additional compute can improve mathematical reasoning through longer attempts or repeated sampling. Prior work finds benefits from both approaches, with their relative effectiveness varying across models, problems, and inference procedures~\citep{snell2025scaling,wu2025inference,hasan2026autofyn}. Long reasoning trajectories can complete computations that short attempts cannot~\citep{mirtaheri2025long}, while broader search exposes alternatives missed by one trajectory~\citep{wang2023selfconsistency,wang2025plansearch}. Can separate measurements predict aggregate scaling under these allocations without observing their outcomes?

We study this question through \emph{strategy discovery} and \emph{conditional execution}. A strategy specifies a proof's load-bearing constructions, claims, and dependencies; execution expands it into a valid proof. A solver could be bottlenecked by either, so two solvers with identical success rates may respond differently to additional compute. We show that short-budget runs and sketch-conditioned completion measurements can predict held-out scaling curves under different compute allocations. We call this framework \emph{Discovery--Execution (DE)}.

In this paper, we approximate the time to strategy acquisition with a geometric distribution, but this is not a requirement for the general DE framework. Under this assumption, when execution is saturated, DE reduces to the \textit{Simple Geometric (SG)} success curve. When execution improves with additional compute, the framework instead predicts different scaling curves and allocation tradeoffs. The prediction of longer reasoning outcomes therefore depends on both strategy acquisition and conditional execution, and cannot always be inferred from short-budget success probability alone.

To test our framework, we introduce \textbf{AOBench (Advanced Olympiad Benchmark)} -- 35 hard, non-geometry problems from 2026 olympiads with human-checked reference proofs, solution sketches of at most 25 words, and explicit load-bearing steps. AOBench is designed to be comparable in size and difficulty to the 30-problem IMO-ProofBench Advanced benchmark~\citep{luong2025towards}. We replicate our study on all 22 non-geometry problems from that benchmark as well. We exclude geometry problems because models can reduce these problems to extensive algebraic calculations whose size makes human audit impractical.


\autoref{fig:framework-overview} summarizes the measurement and prediction setup. A \textit{Parallel-N} compute \textit{allocation} divides a \textit{budget} of $B=8$ compute blocks equally among $N$ independent \textit{attempts}. The \textit{run} succeeds if any attempt produces a correct solution. An \textit{Oracle-Execution} trajectory initializes an attempt with a verified strategy sketch. We measure Parallel-8 and Oracle-Execution success rates to estimate discovery and execution without observing other allocations, and evaluate on Parallel-\textit{N} trajectories for $N=1,2,4$ (\autoref{fig:allocation-model-prediction}). The model-generated proofs are graded for mathematical validity by GPT-5.6 Sol. A stratified sample of 50 proofs was also graded by an International Mathematical Olympiad (IMO) medalist panel to assess agreement with the automated audits.


Our results show when SG predicts aggregate scaling well and when measuring execution improves its predictions. We observe that the GPT models approach execution saturation early, so the framework predicts scaling close to SG, matching the observed curves. In contrast, Claude Opus benefits from additional compute during execution. Incorporating execution with R-DE reduces held-out prediction RMSE from 7.16 to 2.87 solved trials at $N=1$ and from 9.16 to 2.96 at $N=2$. A joint target-trial bootstrap supports the improvement across these allocations (Appendix~\ref{app:prediction-uncertainty}). Even within Opus, the same contrast holds: SG predictions fit well on problems with saturated first-block execution, while incorporating execution improves predictions on the unsaturated subset (\autoref{tab:execution-regimes}).

As a secondary application, these measurements inform whether to extend an unfinished trajectory or start over. R-DE achieves modestly lower average measured regret than each model's retrospectively selected best fixed action, with the largest average gains at early budgets (\autoref{fig:checkpoint-regret}).

In summary, our contributions are:

\begin{itemize}
\item A predictive discovery--execution framework that expresses test-time mathematical reasoning as a convolution of strategy discovery and execution.
\item Predictions of aggregate unaided scaling from separate measurements, testing when execution adds predictive information, with continue-or-restart decisions as an application.
\item AOBench: 35 hard Olympiad problems with human-checked proofs and verified strategy sketches, evaluated alongside IMO-ProofBench Advanced across four models.
\end{itemize}

\section{Related work}

\paragraph{Forecasting inference scaling.} Repeated independent sampling can substantially increase problem coverage~\citep{brown2024monkeys}. \citet{schaeffer2025power} explain how heterogeneity in per-problem success probabilities connects exponential per-problem scaling to aggregate power laws and enables forecasting. \citet{kazdan2025passk} develop a beta-binomial approach to predicting pass@$k$ from limited samples. These works forecast repeated-sampling performance from the distribution of single-attempt success probabilities. We ask when short-attempt success also predicts sequentially extended reasoning, and whether separately measured conditional execution adds predictive information.

\paragraph{Depth versus breadth.} Test-time scaling can deepen one trajectory through self-critique, revision, or budget forcing~\citep{madaan2023selfrefine,muennighoff2025s1}, or broaden search through self-consistency and verifier-selected best-of-$N$~\citep{wang2023selfconsistency,snell2025scaling,wu2025inference}. Long chains can dominate parallel short chains in constructed settings~\citep{mirtaheri2025long}, yet the compute-optimal allocation varies with difficulty~\citep{snell2025scaling,wu2025inference}; learned exploration can extend scaling~\citep{setlur2025e3}, while continued reasoning can plateau or regress~\citep{ghosal2025mirage}. \citet{gu2026gap} find evidence that reduced exploration contributes to sequential sampling's disadvantage relative to parallel sampling in their settings. Reset-and-Discard reallocates repeated attempts across problems to improve coverage under a shared budget~\citep{meir2026red}; our continue-or-restart application instead compares extending and restarting an attempt on the same problem.

\paragraph{Strategy generation and execution.} PlanSearch diversifies natural-language plans before implementation and analyzes success conditioned on an idea, including benefits from short sketches~\citep{wang2025plansearch}. TTS-Uniform distributes inference across explicitly extracted strategies to mitigate selection bias~\citep{wu2025uniform}. In mathematical reasoning, \citet{liang2026strategy} distinguish strategy occurrence from model-relative executability, while \citet{liang2025decoupling} separate high-level reasoning from formal proof generation. Targeted guidance can also unlock unaided failures in mathematics~\citep{agrawal2024hint} and olympiad programming~\citep{shi2024usaco}. In our work, oracle guidance was only used as an intervention to estimate compute-dependent execution, and then predict unaided allocations whose outcomes are not used for estimation.

\paragraph{Olympiad proofs and evaluation.} \citet{huang2025winning} study a model-agnostic verification-and-refinement pipeline for IMO proofs. \citet{luong2025towards} introduce IMO-Bench, including IMO-ProofBench for proof generation and IMO-GradingBench for evaluating proof grading. We use the non-geometry IMO-ProofBench Advanced problems alongside AOBench to evaluate predictions of unaided proof success as compute increases.

\section{Frameworks for Test-Time Scaling}
\label{sec:allocation-model}

\newcounter{proposition}

Let us fix a problem, model, and inference procedure, and let $B\in\mathbb{N}$ denote the available test-time compute budget, measured in discrete compute blocks (fixed token budgets in our experiments). We suppress the problem index in this section; a subscript $n$ elsewhere identifies problem-specific quantities.

\subsection{Simple Geometric (SG) Framework}

Let $q$ denote the probability that a fresh attempt solves the problem within one compute block. A simple extrapolation assumes that each additional block has success probability $q$, conditional on the trajectory remaining unsolved. The resulting success curve after $K$ blocks is geometric:
\begin{equation}
s^{\mathrm{SG}}(K)=1-(1-q)^K.
\label{eq:geometric-extrapolation}
\end{equation}
\autoref{eq:geometric-extrapolation} treats every new block as a fresh chance of success. However, an unfinished trajectory may already contain a viable strategy and need more compute to expand it into a rigorous proof. The SG framework does not account for this progress, so its predictions may be miscalibrated when the compute requirements increase. This motivates us to separate strategy acquisition from proof completion.

\subsection{Discovery--Execution (DE) Framework}

We model proof completion as strategy acquisition followed by execution. Let $A$ denote the block in which a viable strategy is first acquired, and let $E$ denote the number of blocks from acquisition to proof completion, counting the acquisition block as the first. We assume that, conditional on strategy acquisition, the execution does not depend on when the strategy was acquired. \textbf{(Assumption 1)}

Let
\[
p(k)=\Pr(A=k),\qquad \varepsilon(\ell)=\Pr(E\leq\ell\mid A=k),
\]
denote the first-discovery distribution and cumulative execution probability, respectively. By Assumption 1, the conditional probability defining $\varepsilon(\ell)$ is the same for every $k$ with $p(k)>0$. When multiple viable strategies exist, $\varepsilon(\ell)$ represents the discovery-weighted average of their execution probabilities. A strategy acquired in block $k$ has $K-k+1$ blocks available for completion by block $K$. Therefore, the probability of solving within $K$ blocks is
\begin{equation}
s(K)
=(p * \varepsilon)(K)
=\sum_{k=1}^{K}\underbrace{p(k)}_{\text{discovery}}\underbrace{\varepsilon(K-k+1)}_{\text{execution}}.
\label{eq:discovery-execution-convolution}
\end{equation}
The unaided success curve is thus a discrete convolution of strategy discovery and conditional execution. If an oracle strategy is provided in the initial prompt, then $p^{\mathrm{oracle}}(1)=1$ and $p^{\mathrm{oracle}}(k)=0$ for $k>1$. The success curve therefore measures conditional oracle execution:
\begin{equation}
    s^{\mathrm{oracle}}(K)=\varepsilon^{\mathrm{oracle}}(K).
\end{equation}

Assuming $\varepsilon^{\mathrm{oracle}}(\ell)$ approximates $\varepsilon(\ell)$ following natural strategy acquisition, we may use the oracle-execution measurement in the convolution to predict unaided success. \textbf{(Assumption 2)}

\subsection{Geometric Distribution for Discovery}

We now simplify the DE framework by assuming that \textit{strategy acquisition} is memoryless, with probability $\alpha\in(0,1]$ of first viable strategy acquisition. \textbf{(Assumption 3)} We do not impose a geometric form on execution. Under this geometric acquisition assumption,
\begin{equation}
p(k)=\alpha(1-\alpha)^{k-1},\qquad s(K)=\sum_{k=1}^{K}\alpha(1-\alpha)^{k-1}\varepsilon(K-k+1).
\label{eq:geometric-discovery-execution}
\end{equation}
At the first block, fresh proof success has probability $q=s(1)=\alpha\cdot\varepsilon(1)$, which need not equal the acquisition probability. Also, $q$ alone cannot determine the longer-trajectory curve under DE since the execution term also influences it.

We examine the robustness of assumptions 1-3 in \autoref{app:robustness}.

\subsection{Optimal Compute Allocation}
For a total budget $B\in\mathbb N$, assigning $K$ blocks to each arm permits $N=\lfloor B/K\rfloor$ independent attempts. Among these equal-depth independent allocations, the optimal depth $K^\star$ maximizes the probability that at least one attempt solves the problem, conditional on the problem parameters:
\begin{equation}
S_{B}(K)=1-(1-s(K))^{\lfloor B/K\rfloor},\qquad K^\star\in\underset{K\in\{1,\ldots,B\}}{\arg\max}\ S_{B}(K).
\label{eq:allocation-argmax}
\end{equation}

\refstepcounter{proposition}\label{proposition:1}
\paragraph{Proposition \theproposition\ (saturated execution).} Under geometric discovery, suppose execution is saturated within one block, $\varepsilon(1)=1$. Then $q=\alpha$ and $s(K)=1-(1-q)^K$. Consequently, all allocations of $N$ independent attempts of $K$ blocks each that use the full budget $NK=B$ have the same success probability:
\begin{equation}
1-(1-s(K))^N=1-(1-q)^{NK}=1-(1-q)^B.
\label{eq:saturated-allocation-equality}
\end{equation}

Thus, the SG framework is a special case of DE under geometric discovery and saturated first-block execution. When execution is unsaturated, additional compute can also help complete an acquired strategy. The following proposition identifies a regime in which this favors longer trajectories.

\refstepcounter{proposition}\label{proposition:2}
\paragraph{Proposition \theproposition\ (unsaturated execution).} Suppose $\varepsilon(1)<\varepsilon(\ell)$ for some $\ell\in\{2,\ldots,B\}$. Under geometric discovery, as $\alpha\to0$, Eq.~\ref{eq:allocation-argmax} is uniquely maximized to first order in $\alpha$ by $K^\star=B$ and $N^\star=1$.

In the rare-discovery limit, splitting a budget provides no first-order increase in discovery opportunities. A long trajectory leaves more compute to execute an early discovery. Proofs are provided in \autoref{app:propositions}.

\subsection{Parameter Estimation}
\label{sec:joint-estimation}

We estimate fresh success $\widehat q$ from Parallel-8 and conditional completion $\widehat\varepsilon^{\mathrm{oracle}}(k)$ from Oracle-Execution. Section~\ref{sec:experiments} describes the measurement runs.

\paragraph{Discovery--Execution (DE).} Under assumption 2, $\varepsilon^{\mathrm{oracle}}(k)\approx\varepsilon(k)$ for $1\leq k\leq B$, we estimate discovery by correcting fresh success for first-block oracle completion:
\begin{equation}
\widehat{\alpha}=\min\!\left\{1,\frac{\widehat q}{\widehat\varepsilon^{\mathrm{oracle}}(1)}\right\},\qquad \widehat\varepsilon^{\mathrm{oracle}}(1)>0.
\label{eq:de-acquisition-estimator}
\end{equation}
When the empirical fresh rate exceeds first-block oracle completion, the constrained fit sets $\widehat\alpha=1$ and jointly adjusts the execution estimate.

\paragraph{Regularized Discovery--Execution (R-DE).} For $1\leq k\leq B$, let $\pi^{(k)}$ denote the probability of first oracle completion in block $k$, and let $\pi^{(B+1)}$ denote the probability of no completion by block $B$. Then, 
\[
\varepsilon(\ell)=\sum_{k=1}^{\ell}\pi^{(k)},\quad 1\leq\ell\leq B
\]
Let $\boldsymbol\pi=\left(\pi^{(1)},\ldots,\pi^{(B+1)}\right)$. To stabilize estimates from few trials, we use the priors
\[
\alpha\sim\operatorname{Beta}(a,b),\qquad
\boldsymbol\pi\sim\operatorname{Dirichlet}(\boldsymbol d),
\]
with shared parameters $(a,b,\boldsymbol d)$ fitted across all 57 problems separately for each LLM, using only Parallel-8 and Oracle-Execution data. More details about parameter estimation are provided in \autoref{app:joint-estimator}.

\subsection{Continue or Restart}
\label{sec:continue-restart}

\autoref{proposition:1} shows that Simple Geometric assigns the same success probability to continuing an unfinished attempt for $h$ more blocks or starting a fresh $h$-block attempt. Both DE and R-DE can distinguish these actions through the measured execution curve.

For a given problem, continuation contributes an unconditional success increment $s(k+h)-s(k)$, whereas restarting unfinished trajectories contributes $(1-s(k))s(h)$. We compare the expected unconditional gains:
\begin{equation}
\Delta^{(k)}(h)=\mathbb E\!\left[s(k+h)-s(k)-(1-s(k))s(h)\right].
\label{eq:continue-restart-decision}
\end{equation}
A positive expected advantage favors continuation, giving the per-problem policy:
\[
\rho^{(k)}(h)=
\begin{cases}
\text{continue}, & \Delta^{(k)}(h)>0,\\
\text{restart}, & \text{otherwise}.
\end{cases}
\]

\section{Experiments}
\label{sec:experiments}

\subsection{Datasets and Models}

\paragraph{Datasets.} AOBench contains 35 non-geometry problems from twelve olympiads and selection tests held in 2026: 9 algebra, 13 combinatorics, and 13 number theory. The set was fixed before model evaluation, and every problem has a human-audited reference proof, a verified strategy sketch, and three explicit load-bearing steps. We repeat the study on all 22 non-geometry problems from IMO-ProofBench Advanced: 8 algebra, 8 combinatorics, and 6 number theory~\citep{luong2025towards}. \autoref{app:dataset-subtopics} provides details and the AOBench subtopic breakdown.

\paragraph{Models.} We evaluate Muse Spark~1.2, GPT-5.4, GPT-5.5, and Claude Opus~4.8. We report results separately for each model. All models use the same harness and tools, with requests routed to their respective provider APIs.

\subsection{Harness and compute}

Each attempt uses an isolated Docker container and model session. Solvers receive the problem and condition-specific prompt, can execute local code, and cannot access reference solutions, other runs, web tools, or other agents. Networking is restricted to provider endpoints (Appendix~\ref{app:harness}).

A compute block ($1\times$) is a budget of 200k provider-reported output tokens, including reasoning and generated tool calls. Sequential trajectories alternate critique and revision in one persistent Self-Refine session~\citep{madaan2023selfrefine}, up to $8\times=1.6$M tokens. We evaluate the last complete proof at each cumulative block boundary. Early termination is allowed after a minimum of ten rounds and if two consecutive self-critiques find no gap in the generated proof.

\subsection{Measurements}
\label{sec:measurement-arms}

\paragraph{Parallel-$N$ ($N\in\{1,2,4,8\}$).} The total budget is $B=8$ blocks. Parallel-$N$ allocates $N$ independent attempts, each running a fresh trajectory for $K=B/N$ blocks.

\paragraph{Oracle-Execution ($N=1,K=8$).} An eight-block trajectory receives a verified sketch of one reference strategy, limited to 25 words. 

We use three runs of \textbf{Parallel-8 and Oracle-Execution measurements} to estimate our parameters $q$ and $\varepsilon$. From Parallel-8, three runs of 8 fresh attempts give 24 one-block proof outcomes per problem. Their solve fraction gives the fresh success estimate $\widehat q$. From Oracle-Execution, the fraction solved by block $1\leq k\leq B$ estimates $\varepsilon^{\mathrm{oracle}}(k)$.

Parallel-$N$ runs with $N\neq8$ are \textbf{unaided target runs}, and their outcomes are never used to fit our framework parameters. We evaluate $N=1,2$ for all four models and $N=4$ for Muse and GPT-5.5.

\paragraph{Continue or Restart.} At checkpoints $k=1,\ldots,7$, we allocate one additional block ($h=1$) to continuation if R-DE's predicted $\Delta_n^{(k)}(1)>0$ in \autoref{eq:continue-restart-decision}, and to restart otherwise. We evaluate regret (\autoref{eq:regret}) using observed continuations and estimated restart outcomes. Restart outcomes are estimated from six Parallel-2 trajectories per problem whose outcomes are not used to fit any predictor.

\subsection{Alternative Predictors}
\label{sec:comparison-frameworks}
We define three alternative predictors as follows. All predictors use only measurement-run outcomes.

\paragraph{SG.} The saturated-execution special case predicts scaling from fresh success alone. For $u$ solved attempts among $m=24$ Parallel-8 attempts (3 seeds for each run), the finite-bank estimate is
\[
\widehat s^{\rm SG}(K)=1-\frac{\binom{m-u}{K}}{\binom{m}{K}}.
\]
The numerator is zero when $m-u<K$.

\paragraph{R-SG.} To test whether regularization alone explains the gains, we place a shared $\mathrm{Beta}(a,b)$ prior on fresh success $q$. For each model, we fit $a,b$ by maximizing the beta-binomial marginal likelihood of its 57 problems' Parallel-8 counts.

\paragraph{SCT.} Sketch Curve Transfer adds the sketch-conditioned curve's improvement beyond the first block to fresh success, scaled by the initially unsolved fraction.
\[
\widehat s^{\rm SCT}(K)=\widehat q+(1-\widehat q)\bigl[\widehat\varepsilon^{\rm oracle}(K)-\widehat\varepsilon^{\rm oracle}(1)\bigr].
\]

SCT tests whether the unaided curve can be predicted by translating the sketch-conditioned curve to fresh success.

\subsection{Evaluation}

Model-generated proofs are graded by GPT-5.6 Sol using a near-binary rubric with only four allowed scores: 7 for a complete, rigorous proof; 6 or 5 for an otherwise correct proof with exactly one or two minor local defects, respectively; and 0 otherwise. Scores 5--7 count as solved. An IMO medalist panel independently audited 50 distinct proofs and agreed with automated judgments on solved status in 48/50 cases (96.0\%) and reference-step presence in 140/150 cases (93.3\%). \autoref{app:human-validation} gives the protocol and detailed results.

Our primary target is the aggregate cumulative solved-trial count across problems at each compute checkpoint. We report RMSE between predicted and observed counts, with 171 trials per complete curve (lower is better). The Average row reports the square root of the mean model-specific squared RMSE. Execution-subset comparisons use the same calculation on each subset.

For continue-or-restart decisions, let $C$ and $R$ be the observed continuation and estimated restart counts of additional solved trials, and let $V(\rho^{(k)}(h))$ be the count under the selected action. Measured regret is
\begin{equation}
g=\max\{C,R\}-V\!\left(\rho^{(k)}(h)\right).
\label{eq:regret}
\end{equation}

We sum regret across problems and average over checkpoints and then equally over models. We compare R-DE with always continuing, always restarting, and best fixed, which retrospectively selects each model's better constant action. Lower regret means fewer additional successes lost relative to the better observed action on each problem.

\section{Results}
\label{sec:results}

\subsection{When does execution matter?}
\label{sec:allocation-regimes}

\begin{figure}[htbp]
\centering
\definecolor{plotunaided}{HTML}{505050}
\definecolor{plotorange}{HTML}{D97706}
\definecolor{plotrde}{HTML}{D94B40}
\definecolor{plottick}{HTML}{44505A}
\definecolor{plotgroup}{HTML}{17212B}
\definecolor{plotaxis}{HTML}{8C959D}
\definecolor{plotgrid}{HTML}{D7DCDF}

\pgfplotsset{ compact scaling panel/.style={ scale only axis, width=0.204\linewidth, height=2.35cm, xmin=0.7, xmax=8.3, ymin=0, ymax=171, xtick={2,4,6,8}, xticklabels={$2\times$,$4\times$,$6\times$,$8\times$}, ytick={20,50,80,110,140,171}, tick label style={font=\normalfont\normalsize,text=black}, xticklabel style={font=\normalfont\normalsize,text=black}, label style={font=\normalfont\normalsize,text=black}, axis line style={draw=plotaxis,line width=0.7pt}, axis x line*=bottom, axis y line*=left, tick style={draw=none}, ymajorgrids=true, grid style={draw=plotgrid,opacity=0.8,line width=0.5pt}, clip=false,
  }, compact unaided curve/.style={ draw=plotunaided, line width=1pt, mark=*, mark size=1.5pt, mark options={solid,fill=plotunaided,draw=plotunaided,line width=0.25pt},
  }, compact oracle curve/.style={ draw=plotorange, line width=1pt, mark=*, mark size=1.5pt, mark options={solid,fill=plotorange,draw=plotorange!65!black,line width=0.25pt},
  }, compact rde curve/.style={ draw=plotrde, dash pattern=on 6pt off 3pt, line width=1.6pt, no marks,
  },
}

\begin{tikzpicture}[font=\normalfont\normalsize,text=black]
\begin{groupplot}[ group style={group size=4 by 1,horizontal sep=0.38cm}, compact scaling panel, xticklabels=\empty,
]
\nextgroupplot
  \addplot[draw=none,name path=muse-unaided] coordinates {(1,56) (2,63) (3,64) (4,64) (5,64) (6,64) (7,64) (8,68)};
  \addplot[draw=none,name path=muse-oracle] coordinates {(1,102) (2,107) (3,108) (4,109) (5,110) (6,110) (7,110) (8,111)};
  \addplot[fill=plotorange,fill opacity=0.13,draw=none] fill between[of=muse-unaided and muse-oracle];
  \addplot[compact unaided curve] coordinates {(1,56) (2,63) (3,64) (4,64) (5,64) (6,64) (7,64) (8,68)};
  \addplot[compact oracle curve] coordinates {(1,102) (2,107) (3,108) (4,109) (5,110) (6,110) (7,110) (8,111)};
  \addplot[compact rde curve] coordinates {(1,46.055945) (2,56.307593) (3,61.661585) (4,65.614640) (5,68.689334) (6,70.553485) (7,72.006720) (8,73.628419)};
  \node[font=\normalfont\small,text=plotorange,align=center,inner sep=1pt] at (axis cs:4.5,87) {$\leq25$-word sketch};

\nextgroupplot[yticklabels=\empty]
  \addplot[draw=none,name path=gptf-unaided] coordinates {(1,84) (2,103) (3,109) (4,111) (5,116) (6,117) (7,121) (8,122)};
  \addplot[draw=none,name path=gptf-oracle] coordinates {(1,139) (2,148) (3,152) (4,153) (5,153) (6,154) (7,154) (8,154)};
  \addplot[fill=plotorange,fill opacity=0.13,draw=none] fill between[of=gptf-unaided and gptf-oracle];
  \addplot[compact unaided curve] coordinates {(1,84) (2,103) (3,109) (4,111) (5,116) (6,117) (7,121) (8,122)};
  \addplot[compact oracle curve] coordinates {(1,139) (2,148) (3,152) (4,153) (5,153) (6,154) (7,154) (8,154)};
  \addplot[compact rde curve] coordinates {(1,81.153547) (2,100.466019) (3,110.704497) (4,116.392746) (5,119.798609) (6,122.628464) (7,124.626002) (8,126.215350)};

\nextgroupplot[yticklabels=\empty]
  \addplot[draw=none,name path=gptv-unaided] coordinates {(1,138) (2,147) (3,153) (4,155) (5,157) (6,158) (7,158) (8,158)};
  \addplot[draw=none,name path=gptv-oracle] coordinates {(1,162) (2,167) (3,167) (4,167) (5,167) (6,168) (7,168) (8,168)};
  \addplot[fill=plotorange,fill opacity=0.13,draw=none] fill between[of=gptv-unaided and gptv-oracle];
  \addplot[compact unaided curve] coordinates {(1,138) (2,147) (3,153) (4,155) (5,157) (6,158) (7,158) (8,158)};
  \addplot[compact oracle curve] coordinates {(1,162) (2,167) (3,167) (4,167) (5,167) (6,168) (7,168) (8,168)};
  \addplot[compact rde curve] coordinates {(1,137.739816) (2,147.002792) (3,149.390011) (4,150.830279) (5,151.848791) (6,153.334427) (7,153.988371) (8,154.525772)};
\nextgroupplot[yticklabels=\empty]
  \addplot[draw=none,name path=opus-unaided] coordinates {(1,91) (2,115) (3,121) (4,124) (5,125) (6,127) (7,129) (8,129)};
  \addplot[draw=none,name path=opus-oracle] coordinates {(1,113) (2,136) (3,142) (4,146) (5,150) (6,150) (7,152) (8,153)};
  \addplot[fill=plotorange,fill opacity=0.13,draw=none] fill between[of=opus-unaided and opus-oracle];
  \addplot[compact unaided curve] coordinates {(1,91) (2,115) (3,121) (4,124) (5,125) (6,127) (7,129) (8,129)};
  \addplot[compact oracle curve] coordinates {(1,113) (2,136) (3,142) (4,146) (5,150) (6,150) (7,152) (8,153)};
  \addplot[compact rde curve] coordinates {(1,88.010824) (2,109.638981) (3,117.654136) (4,121.935847) (5,126.793777) (6,127.984053) (7,130.222508) (8,131.653538)};

\end{groupplot}

\draw[plotunaided,line width=1pt] ([xshift=-40mm,yshift=5mm]{$(group c1r1.north)!0.5!(group c4r1.north)$}) -- ++(5mm,0);
\node[font=\normalfont\normalsize,anchor=west] at ([xshift=-34mm,yshift=5mm]{$(group c1r1.north)!0.5!(group c4r1.north)$}) {Unaided};
\draw[plotorange,line width=1pt] ([xshift=-14mm,yshift=5mm]{$(group c1r1.north)!0.5!(group c4r1.north)$}) -- ++(5mm,0);
\node[font=\normalfont\normalsize,anchor=west] at ([xshift=-8mm,yshift=5mm]{$(group c1r1.north)!0.5!(group c4r1.north)$}) {Sketch};
\draw[plotrde,dash pattern=on 6pt off 3pt,line width=1.6pt] ([xshift=24mm,yshift=5mm]{$(group c1r1.north)!0.5!(group c4r1.north)$}) -- ++(5mm,0);
\node[font=\normalfont\normalsize,anchor=west] at ([xshift=30mm,yshift=5mm]{$(group c1r1.north)!0.5!(group c4r1.north)$}) {R-DE};

\node[font=\normalfont\normalsize,rotate=90,anchor=south] at ([xshift=-7mm]group c1r1.west) {Solved};
\end{tikzpicture}
\par\vspace{6mm}
\begingroup
\definecolor{errorogt}{HTML}{249447}
\colorlet{errorrsg}{cyan!80!blue}
\definecolor{errorgeo}{HTML}{2077B4}
\definecolor{errorde}{HTML}{8246C5}
\definecolor{errorrde}{HTML}{D94B40}
\definecolor{errortick}{HTML}{44505A}
\definecolor{errorgroup}{HTML}{17212B}
\definecolor{erroraxis}{HTML}{8C959D}
\definecolor{errorgrid}{HTML}{D7DCDF}
\begin{tikzpicture}[font=\normalfont\normalsize,text=black]
\begin{groupplot}[group style={group size=4 by 1,horizontal sep=0.38cm},scale only axis,width=0.204\linewidth,height=2.35cm,xmin=0.7,xmax=8.3,ymin=-15,ymax=30,xtick={2,4,6,8},xticklabels={$2\times$,$4\times$,$6\times$,$8\times$},ytick={-10,0,10,20,30},tick label style={font=\normalfont\normalsize,text=black},xticklabel style={font=\normalfont\normalsize,text=black},label style={font=\normalfont\normalsize,text=black},axis line style={draw=erroraxis,line width=0.7pt},axis x line*=bottom,axis y line*=left,tick style={draw=none},ymajorgrids=true,grid style={draw=errorgrid,opacity=0.8,line width=0.5pt},clip=true]
\nextgroupplot[]
\addplot[black!65,solid,line width=1.2pt,no marks] coordinates {(0.7,0) (8.3,0)};
\addplot[draw=errorogt,solid,line width=1pt,mark=square*,mark size=1.3pt,mark options={solid,fill=errorogt,draw=errorogt!65!black,line width=0.25pt}] coordinates {(1,9.875000) (2,12.041667) (3,12.083333) (4,11.291667) (5,10.583333) (6,10.583333) (7,10.583333) (8,14.458333)};
\addplot[draw=errorgeo,solid,line width=1pt,mark=*,mark size=1.5pt,mark options={solid,fill=errorgeo,draw=errorgeo!65!black,line width=0.25pt}] coordinates {(1,9.875000) (2,5.467391) (3,0.155138) (4,-4.199322) (5,-7.467109) (6,-10.035915) (7,-12.124584) (8,-9.871576)};
\addplot[draw=errorrsg,solid,line width=1pt,mark=pentagon*,mark size=1.7pt,mark options={solid,fill=errorrsg,draw=errorrsg!65!black,line width=0.25pt}] coordinates {(1,9.876403) (2,5.417069) (3,0.091916) (4,-4.243092) (5,-7.505194) (6,-10.093844) (7,-12.222957) (8,-10.021226)};
\addplot[draw=errorde,solid,line width=1pt,mark=triangle*,mark size=2.1pt,mark options={solid,fill=errorde,draw=errorde!65!black,line width=0.25pt}] coordinates {(1,10.236111) (2,5.599826) (3,1.046857) (4,-2.539975) (5,-5.635761) (6,-7.340630) (7,-8.660222) (8,-6.173030)};
\addplot[draw=errorrde,solid,line width=1pt,mark=diamond*,mark size=1.9pt,mark options={solid,fill=errorrde,draw=errorrde!65!black,line width=0.25pt}] coordinates {(1,9.944055) (2,6.692407) (3,2.338415) (4,-1.614640) (5,-4.689334) (6,-6.553485) (7,-8.006720) (8,-5.628419)};
\nextgroupplot[yticklabels=\empty]
\addplot[black!65,solid,line width=1.2pt,no marks] coordinates {(0.7,0) (8.3,0)};
\addplot[draw=errorogt,solid,line width=1pt,mark=square*,mark size=1.3pt,mark options={solid,fill=errorogt,draw=errorogt!65!black,line width=0.25pt}] coordinates {(1,3.125000) (2,14.291667) (3,16.750000) (4,17.833333) (5,22.833333) (6,22.833333) (7,26.833333) (8,27.833333)};
\addplot[draw=errorgeo,solid,line width=1pt,mark=*,mark size=1.5pt,mark options={solid,fill=errorgeo,draw=errorgeo!65!black,line width=0.25pt}] coordinates {(1,3.125000) (2,3.032609) (3,-0.178360) (4,-4.007905) (5,-3.261858) (6,-5.592358) (7,-4.289586) (8,-5.509539)};
\addplot[draw=errorrsg,solid,line width=1pt,mark=pentagon*,mark size=1.7pt,mark options={solid,fill=errorrsg,draw=errorrsg!65!black,line width=0.25pt}] coordinates {(1,3.181667) (2,3.112638) (3,-0.219833) (4,-4.108002) (5,-3.337030) (6,-5.602179) (7,-4.235674) (8,-5.421499)};
\addplot[draw=errorde,solid,line width=1pt,mark=triangle*,mark size=2.1pt,mark options={solid,fill=errorde,draw=errorde!65!black,line width=0.25pt}] coordinates {(1,3.291667) (2,2.557002) (3,-2.343768) (4,-6.311079) (5,-4.541160) (6,-5.820907) (7,-3.530974) (8,-3.863989)};
\addplot[draw=errorrde,solid,line width=1pt,mark=diamond*,mark size=1.9pt,mark options={solid,fill=errorrde,draw=errorrde!65!black,line width=0.25pt}] coordinates {(1,2.846453) (2,2.533981) (3,-1.704497) (4,-5.392746) (5,-3.798609) (6,-5.628464) (7,-3.626002) (8,-4.215350)};
\nextgroupplot[yticklabels=\empty]
\addplot[black!65,solid,line width=1.2pt,no marks] coordinates {(0.7,0) (8.3,0)};
\addplot[draw=errorogt,solid,line width=1pt,mark=square*,mark size=1.3pt,mark options={solid,fill=errorogt,draw=errorogt!65!black,line width=0.25pt}] coordinates {(1,0.250000) (2,6.125000) (3,12.125000) (4,14.125000) (5,16.125000) (6,17.125000) (7,17.125000) (8,17.125000)};
\addplot[draw=errorgeo,solid,line width=1pt,mark=*,mark size=1.5pt,mark options={solid,fill=errorgeo,draw=errorgeo!65!black,line width=0.25pt}] coordinates {(1,0.250000) (2,-0.532609) (3,2.181818) (4,2.434783) (5,3.284020) (6,3.430481) (7,2.749809) (8,2.182340)};
\addplot[draw=errorrsg,solid,line width=1pt,mark=pentagon*,mark size=1.7pt,mark options={solid,fill=errorrsg,draw=errorrsg!65!black,line width=0.25pt}] coordinates {(1,0.374491) (2,-0.531378) (3,1.907663) (4,1.963021) (5,2.665683) (6,2.699899) (7,1.931966) (8,1.294904)};
\addplot[draw=errorde,solid,line width=1pt,mark=triangle*,mark size=2.1pt,mark options={solid,fill=errorde,draw=errorde!65!black,line width=0.25pt}] coordinates {(1,0.513889) (2,-0.519097) (3,2.660807) (4,3.115370) (5,4.076484) (6,4.189395) (7,3.574743) (8,3.070785)};
\addplot[draw=errorrde,solid,line width=1pt,mark=diamond*,mark size=1.9pt,mark options={solid,fill=errorrde,draw=errorrde!65!black,line width=0.25pt}] coordinates {(1,0.260184) (2,-0.002792) (3,3.609989) (4,4.169721) (5,5.151209) (6,4.665573) (7,4.011629) (8,3.474228)};
\nextgroupplot[yticklabels=\empty]
\addplot[black!65,solid,line width=1.2pt,no marks] coordinates {(0.7,0) (8.3,0)};
\addplot[draw=errorogt,solid,line width=1pt,mark=square*,mark size=1.3pt,mark options={solid,fill=errorogt,draw=errorogt!65!black,line width=0.25pt}] coordinates {(1,2.750000) (2,11.416667) (3,13.375000) (4,13.375000) (5,11.166667) (6,13.166667) (7,13.166667) (8,12.166667)};
\addplot[draw=errorgeo,solid,line width=1pt,mark=*,mark size=1.5pt,mark options={solid,fill=errorgeo,draw=errorgeo!65!black,line width=0.25pt}] coordinates {(1,2.750000) (2,11.163043) (3,9.544960) (4,7.922643) (5,5.822205) (6,5.620494) (7,5.988238) (8,4.731368)};
\addplot[draw=errorrsg,solid,line width=1pt,mark=pentagon*,mark size=1.7pt,mark options={solid,fill=errorrsg,draw=errorrsg!65!black,line width=0.25pt}] coordinates {(1,2.765140) (2,11.358885) (3,9.800514) (4,8.135214) (5,5.915464) (6,5.536020) (7,5.690850) (8,4.206544)};
\addplot[draw=errorde,solid,line width=1pt,mark=triangle*,mark size=2.1pt,mark options={solid,fill=errorde,draw=errorde!65!black,line width=0.25pt}] coordinates {(1,3.722222) (2,6.084201) (3,4.888401) (4,4.425864) (5,1.044315) (6,1.619882) (7,2.608416) (8,1.874891)};
\addplot[draw=errorrde,solid,line width=1pt,mark=diamond*,mark size=1.9pt,mark options={solid,fill=errorrde,draw=errorrde!65!black,line width=0.25pt}] coordinates {(1,2.989176) (2,5.361019) (3,3.345864) (4,2.064153) (5,-1.793777) (6,-0.984053) (7,-1.222508) (8,-2.653538)};
\end{groupplot}
\node[font=\normalfont\normalsize,anchor=north] at ([yshift=-3mm]group c1r1.south) {Muse Spark~1.2};
\node[font=\normalfont\normalsize,anchor=north] at ([yshift=-3mm]group c2r1.south) {GPT-5.4};
\node[font=\normalfont\normalsize,anchor=north] at ([yshift=-3mm]group c3r1.south) {GPT-5.5};
\node[font=\normalfont\normalsize,anchor=north] at ([yshift=-3mm]group c4r1.south) {Claude Opus~4.8};
\node[rotate=90,anchor=south,font=\normalfont\normalsize] at ([xshift=-7mm]group c1r1.west) {Solved $-$ Predicted};
\node[anchor=north,font=\normalfont\normalsize] at ([yshift=-7.5mm]{$(group c1r1.south)!0.5!(group c4r1.south)$}) {Inference budget};
\draw[errorogt,solid,line width=1pt] ([xshift=-57mm,yshift=5mm]{$(group c1r1.north)!0.5!(group c4r1.north)$}) -- ++(5mm,0);
\draw[errorogt,mark=square*,mark size=1.3pt,mark options={fill=errorogt,draw=errorogt!65!black,line width=0.25pt}] plot coordinates {([xshift=-54.5mm,yshift=5mm]{$(group c1r1.north)!0.5!(group c4r1.north)$})};
\node[anchor=west,font=\normalfont\normalsize] at ([xshift=-51mm,yshift=5mm]{$(group c1r1.north)!0.5!(group c4r1.north)$}) {SCT};
\draw[errorgeo,solid,line width=1pt] ([xshift=-33mm,yshift=5mm]{$(group c1r1.north)!0.5!(group c4r1.north)$}) -- ++(5mm,0);
\draw[errorgeo,mark=*,mark size=1.5pt,mark options={fill=errorgeo,draw=errorgeo!65!black,line width=0.25pt}] plot coordinates {([xshift=-30.5mm,yshift=5mm]{$(group c1r1.north)!0.5!(group c4r1.north)$})};
\node[anchor=west,font=\normalfont\normalsize] at ([xshift=-27mm,yshift=5mm]{$(group c1r1.north)!0.5!(group c4r1.north)$}) {SG};
\draw[errorrsg,solid,line width=1pt] ([xshift=-12mm,yshift=5mm]{$(group c1r1.north)!0.5!(group c4r1.north)$}) -- ++(5mm,0);
\draw[errorrsg,mark=pentagon*,mark size=1.7pt,mark options={fill=errorrsg,draw=errorrsg!65!black,line width=0.25pt}] plot coordinates {([xshift=-9.5mm,yshift=5mm]{$(group c1r1.north)!0.5!(group c4r1.north)$})};
\node[anchor=west,font=\normalfont\normalsize] at ([xshift=-6mm,yshift=5mm]{$(group c1r1.north)!0.5!(group c4r1.north)$}) {R-SG};
\draw[errorde,solid,line width=1pt] ([xshift=14mm,yshift=5mm]{$(group c1r1.north)!0.5!(group c4r1.north)$}) -- ++(5mm,0);
\draw[errorde,mark=triangle*,mark size=2.1pt,mark options={fill=errorde,draw=errorde!65!black,line width=0.25pt}] plot coordinates {([xshift=16.5mm,yshift=5mm]{$(group c1r1.north)!0.5!(group c4r1.north)$})};
\node[anchor=west,font=\normalfont\normalsize] at ([xshift=20mm,yshift=5mm]{$(group c1r1.north)!0.5!(group c4r1.north)$}) {DE};
\draw[errorrde,solid,line width=1pt] ([xshift=35mm,yshift=5mm]{$(group c1r1.north)!0.5!(group c4r1.north)$}) -- ++(5mm,0);
\draw[errorrde,mark=diamond*,mark size=1.9pt,mark options={fill=errorrde,draw=errorrde!65!black,line width=0.25pt}] plot coordinates {([xshift=37.5mm,yshift=5mm]{$(group c1r1.north)!0.5!(group c4r1.north)$})};
\node[anchor=west,font=\normalfont\normalsize] at ([xshift=41mm,yshift=5mm]{$(group c1r1.north)!0.5!(group c4r1.north)$}) {R-DE};
\end{tikzpicture}
\endgroup
\caption{\textbf{Measured execution and held-out Parallel-1 scaling ($N=1$).} Top: sketch-conditioned and unaided successes, with dashed R-DE predictions. Bottom: difference between observed and predicted solved counts. Each condition has 171 trajectories across 57 problems (3 seeds each).}
\label{fig:headline}
\label{fig:allocation-model-prediction}
\end{figure}
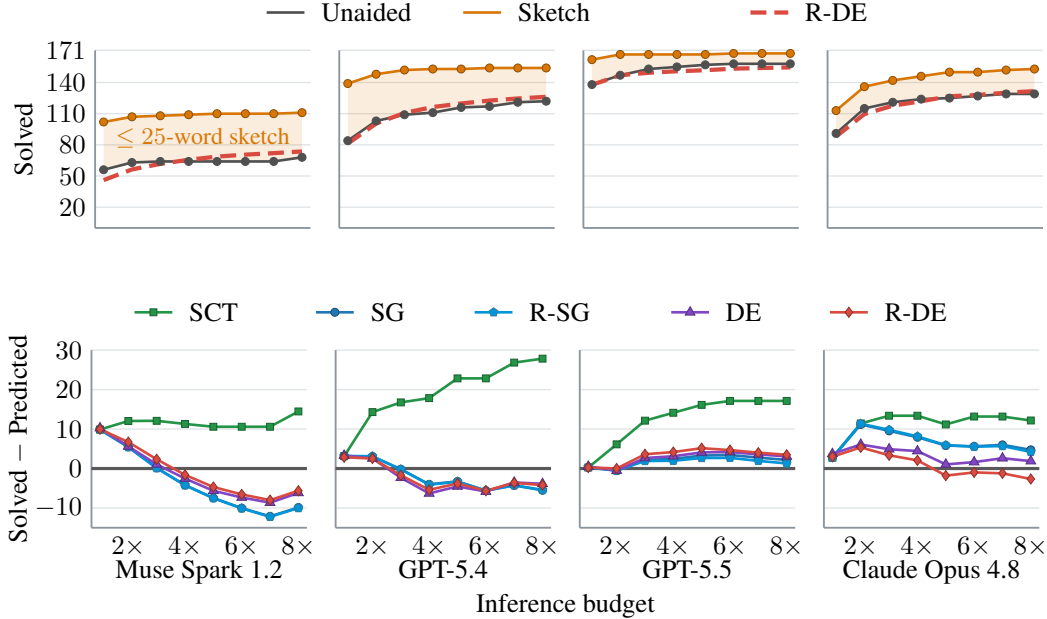

\begin{table}[t]
\caption{Prediction RMSE (solved-trial counts) across 57 problems and 171 trials per model. Bold marks the lowest error within each row and allocation.}
\label{tab:predictor-comparison}
\label{tab:n2-predictor-comparison}
\centering\small\setlength{\tabcolsep}{4pt}
\begin{tabular}{@{}lrrrrr@{\hspace{10pt}}rrrrr@{}}
\toprule & \multicolumn{5}{c}{$N=1$} & \multicolumn{5}{c}{$N=2$} \\
\cmidrule(lr){2-6}\cmidrule(l){7-11}
Model & SG & R-SG & SCT & DE & R-DE & SG & R-SG & SCT & DE & R-DE \\
\midrule
Muse & 8.26 & 8.31 & 11.52 & 6.54 & \textbf{6.25} & 5.61 & \textbf{5.58} & 10.49 & 6.26 & \textbf{5.58} \\
GPT-5.4 & 3.96 & 3.98 & 20.46 & 4.25 & \textbf{3.93} & 2.83 & 2.83 & 9.97 & 2.32 & \textbf{2.16} \\
GPT-5.5 & 2.40 & \textbf{1.86} & 13.80 & 3.04 & 3.66 & 3.59 & \textbf{3.03} & 6.47 & 3.88 & 3.46 \\
Opus & 7.16 & 7.20 & 11.81 & 3.68 & \textbf{2.87} & 9.16 & 9.10 & 5.48 & 6.70 & \textbf{2.96} \\
\midrule
Average & 5.94 & 5.92 & 14.84 & 4.57 & \textbf{4.36} & 5.84 & 5.73 & 8.39 & 5.11 & \textbf{3.76} \\
\bottomrule
\end{tabular}
\end{table}

The predictive value of measuring execution varies across models (Figure~\ref{fig:headline}). GPT-5.5 completes 162/171 oracle trajectories within one block and 168/171 within eight. With little further execution gain (\autoref{proposition:1}), SG closely predicts its unaided scaling (RMSE 2.40), as well as GPT-5.4's (Table~\ref{tab:predictor-comparison}).

Opus's oracle successes rise from 113/171 to 153/171 over eight blocks. Incorporating execution improves its held-out forecasts, reducing SG's RMSE of 7.16 to 3.68 under DE and 2.87 under R-DE. R-SG gives 7.20, so regularization alone does not explain the improvement. Flattening the measured execution curve also worsens Opus's predictions (Appendix~\ref{app:execution-ablations}).

\begin{table}[!htbp]
\caption{Opus prediction RMSE (solved-trial counts), split by first-block execution saturation. Bold marks row minima. DE helps when execution is not saturated.}
\label{tab:execution-regimes}
\centering\small\setlength{\tabcolsep}{4pt}
\begin{tabular}{llrrr}
\toprule $N$ & Execution & Trials & SG & DE \\
\midrule
\multirow{2}{*}{1} & Saturated & 87 & \textbf{2.40} & 2.65 \\
 & Unsaturated & 84 & 5.50 & \textbf{1.86} \\
\midrule
\multirow{2}{*}{2} & Saturated & 87 & \textbf{2.17} & 2.50 \\
 & Unsaturated & 84 & 8.25 & \textbf{5.22} \\
\bottomrule
\end{tabular}
\end{table}

Within Opus, SG predicts well when all three oracle trajectories solve in one block, and DE offers no improvement (Table~\ref{tab:execution-regimes}). On the remaining 28 problems, DE reduces RMSE from 5.50 to 1.86 at $N=1$ and from 8.25 to 5.22 at $N=2$.

\subsection{Predicting across alternate compute allocations}
\label{sec:allocation-transfer}

At $N=2$, R-DE has the lowest across-model RMSE: 3.76 versus 5.84 for SG and 5.73 for R-SG (Table~\ref{tab:predictor-comparison}). For Opus, R-DE reduces RMSE from SG's 9.16 to 2.96, while R-SG remains best for GPT-5.5.

Combining $N=1$ and $N=2$, Opus's RMSE falls from 8.22 under SG and 8.20 under R-SG to 2.91 under R-DE. Both reductions have 95\% paired target-bootstrap percentile intervals above zero (Appendix~\ref{app:prediction-uncertainty}).

For Muse, R-DE closely predicts final $N=2$ success (77.3 versus 77) but underpredicts at $N=4$ (78.5 versus 93). A post-hoc substitution of Parallel-4 first-block outcomes raises the latter prediction to 94.4, pointing to a lack of transfer in first-block success across allocations (Appendix~\ref{app:muse-n4-diagnostics}).

\paragraph{Transfer without test-problem sketches.} Across 50 splits, we learn shared priors from 38 problems and predict individual four-block unaided trajectories on 19 held-out problems using only their Parallel-8 outcomes. For Opus, R-DE reduces mean solved-count RMSE from R-SG's 6.24 to 5.04, or 3.40 with test-problem oracle measurements. Muse also improves, GPT-5.4 is nearly tied, and GPT-5.5 favors R-SG (Appendix~\ref{app:prior-sensitivity}).

\subsection{Choosing whether to continue or restart}
\label{sec:checkpoint-choices}

\begin{table}[!htbp]
\caption{Mean next-block regret (solved trials lost), $k=1,\ldots,7$. Best fixed selects each model's better constant action retrospectively. Bold marks row minima.}
\label{tab:checkpoint-regret}
\centering\small
\begin{tabular}{lrrrr}
\toprule Model & Always continue & Always restart & Best fixed & R-DE \\
\midrule
Muse Spark~1.2 & 9.74 & \textbf{0.57} & \textbf{0.57} & 1.57 \\
GPT-5.4 & 3.14 & 3.62 & 3.14 & \textbf{2.29} \\
GPT-5.5 & 2.10 & 1.19 & 1.19 & \textbf{0.76} \\
Claude Opus~4.8 & 2.24 & 3.14 & 2.24 & \textbf{1.38} \\
\midrule
Average & 4.30 & 2.13 & 1.79 & \textbf{1.50} \\
\bottomrule
\end{tabular}
\end{table}

\begin{figure}[htbp]
\centering
\begingroup
\definecolor{regcontinue}{HTML}{4C91C7}
\definecolor{regrestart}{HTML}{52B567}
\definecolor{regfixed}{HTML}{9A70D6}
\definecolor{regrde}{HTML}{D94B40}
\definecolor{regaxis}{HTML}{8C959D}
\definecolor{reggrid}{HTML}{D7DCDF}
\begin{tikzpicture}[font=\normalfont\normalsize,text=black]
\begin{groupplot}[group style={group size=4 by 1,horizontal sep=0.38cm},scale only axis,width=0.204\linewidth,height=2.35cm,xmin=0.7,xmax=8.3,ymin=0,ymax=15,xtick={2,4,6,8},xticklabels={$2\times$,$4\times$,$6\times$,$8\times$},ytick={0,5,10,15},tick label style={font=\normalfont\normalsize,text=black},label style={font=\normalfont\normalsize,text=black},axis line style={draw=regaxis,line width=0.7pt},axis x line*=bottom,axis y line*=left,tick style={draw=none},ymajorgrids=true,grid style={draw=reggrid,opacity=0.8,line width=0.5pt},clip=true]
\nextgroupplot
\addplot[solid,line width=2pt,draw=regfixed!65,no marks] coordinates {(2,3.50000000) (3,0.00000000) (4,0.00000000) (5,0.00000000) (6,0.00000000) (7,0.00000000) (8,0.50000000)};
\addplot[solid,line width=1pt,draw=regcontinue,mark=*,mark size=1.5pt,mark options={fill=regcontinue,draw=regcontinue!65!black,line width=0.25pt}] coordinates {(2,10.00000000) (3,10.00000000) (4,10.33333333) (5,10.33333333) (6,10.33333333) (7,10.33333333) (8,6.83333333)};
\addplot[solid,line width=1pt,draw=regrestart,mark=square*,mark size=1.35pt,mark options={fill=regrestart,draw=regrestart!65!black,line width=0.25pt}] coordinates {(2,3.50000000) (3,0.00000000) (4,0.00000000) (5,0.00000000) (6,0.00000000) (7,0.00000000) (8,0.50000000)};
\addplot[solid,line width=1pt,draw=regfixed,mark=triangle,mark size=2.0pt,only marks,mark options={fill=white,draw=regfixed,line width=0.6pt}] coordinates {(2,3.50000000) (3,0.00000000) (4,0.00000000) (5,0.00000000) (6,0.00000000) (7,0.00000000) (8,0.50000000)};
\addplot[solid,line width=1pt,draw=regrde,mark=diamond*,mark size=1.9pt,mark options={fill=regrde,draw=regrde!65!black,line width=0.25pt}] coordinates {(2,4.50000000) (3,0.66666667) (4,1.00000000) (5,1.33333333) (6,1.00000000) (7,1.00000000) (8,1.50000000)};
\nextgroupplot[yticklabels=\empty]
\addplot[solid,line width=2pt,draw=regfixed!65,no marks] coordinates {(2,3.66666667) (3,3.66666667) (4,3.66666667) (5,3.00000000) (6,3.00000000) (7,2.66666667) (8,2.33333333)};
\addplot[solid,line width=1pt,draw=regcontinue,mark=*,mark size=1.5pt,mark options={fill=regcontinue,draw=regcontinue!65!black,line width=0.25pt}] coordinates {(2,3.66666667) (3,3.66666667) (4,3.66666667) (5,3.00000000) (6,3.00000000) (7,2.66666667) (8,2.33333333)};
\addplot[solid,line width=1pt,draw=regrestart,mark=square*,mark size=1.35pt,mark options={fill=regrestart,draw=regrestart!65!black,line width=0.25pt}] coordinates {(2,10.00000000) (3,4.66666667) (4,1.33333333) (5,4.00000000) (6,1.00000000) (7,3.66666667) (8,0.66666667)};
\addplot[solid,line width=1pt,draw=regfixed,mark=triangle,mark size=2.0pt,only marks,mark options={fill=white,draw=regfixed,line width=0.6pt}] coordinates {(2,3.66666667) (3,3.66666667) (4,3.66666667) (5,3.00000000) (6,3.00000000) (7,2.66666667) (8,2.33333333)};
\addplot[solid,line width=1pt,draw=regrde,mark=diamond*,mark size=1.9pt,mark options={fill=regrde,draw=regrde!65!black,line width=0.25pt}] coordinates {(2,4.33333333) (3,1.66666667) (4,1.66666667) (5,3.00000000) (6,1.66666667) (7,2.33333333) (8,1.33333333)};
\nextgroupplot[yticklabels=\empty]
\addplot[solid,line width=2pt,draw=regfixed!65,no marks] coordinates {(2,3.83333333) (3,2.16666667) (4,1.16666667) (5,1.00000000) (6,0.16666667) (7,0.00000000) (8,0.00000000)};
\addplot[solid,line width=1pt,draw=regcontinue,mark=*,mark size=1.5pt,mark options={fill=regcontinue,draw=regcontinue!65!black,line width=0.25pt}] coordinates {(2,5.00000000) (3,2.33333333) (4,2.50000000) (5,1.83333333) (6,1.00000000) (7,1.00000000) (8,1.00000000)};
\addplot[solid,line width=1pt,draw=regrestart,mark=square*,mark size=1.35pt,mark options={fill=regrestart,draw=regrestart!65!black,line width=0.25pt}] coordinates {(2,3.83333333) (3,2.16666667) (4,1.16666667) (5,1.00000000) (6,0.16666667) (7,0.00000000) (8,0.00000000)};
\addplot[solid,line width=1pt,draw=regfixed,mark=triangle,mark size=2.0pt,only marks,mark options={fill=white,draw=regfixed,line width=0.6pt}] coordinates {(2,3.83333333) (3,2.16666667) (4,1.16666667) (5,1.00000000) (6,0.16666667) (7,0.00000000) (8,0.00000000)};
\addplot[solid,line width=1pt,draw=regrde,mark=diamond*,mark size=1.9pt,mark options={fill=regrde,draw=regrde!65!black,line width=0.25pt}] coordinates {(2,1.33333333) (3,1.16666667) (4,1.00000000) (5,0.33333333) (6,0.83333333) (7,0.33333333) (8,0.33333333)};
\nextgroupplot[yticklabels=\empty]
\addplot[solid,line width=2pt,draw=regfixed!65,no marks] coordinates {(2,3.33333333) (3,2.00000000) (4,2.50000000) (5,2.50000000) (6,1.83333333) (7,1.66666667) (8,1.83333333)};
\addplot[solid,line width=1pt,draw=regcontinue,mark=*,mark size=1.5pt,mark options={fill=regcontinue,draw=regcontinue!65!black,line width=0.25pt}] coordinates {(2,3.33333333) (3,2.00000000) (4,2.50000000) (5,2.50000000) (6,1.83333333) (7,1.66666667) (8,1.83333333)};
\addplot[solid,line width=1pt,draw=regrestart,mark=square*,mark size=1.35pt,mark options={fill=regrestart,draw=regrestart!65!black,line width=0.25pt}] coordinates {(2,12.16666667) (3,3.50000000) (4,2.33333333) (5,1.00000000) (6,1.33333333) (7,1.66666667) (8,0.00000000)};
\addplot[solid,line width=1pt,draw=regfixed,mark=triangle,mark size=2.0pt,only marks,mark options={fill=white,draw=regfixed,line width=0.6pt}] coordinates {(2,3.33333333) (3,2.00000000) (4,2.50000000) (5,2.50000000) (6,1.83333333) (7,1.66666667) (8,1.83333333)};
\addplot[solid,line width=1pt,draw=regrde,mark=diamond*,mark size=1.9pt,mark options={fill=regrde,draw=regrde!65!black,line width=0.25pt}] coordinates {(2,3.00000000) (3,1.50000000) (4,1.66666667) (5,1.00000000) (6,1.00000000) (7,1.16666667) (8,0.33333333)};
\end{groupplot}
\node[anchor=north,align=center] at ([yshift=-3mm]group c1r1.south) {Muse Spark~1.2};
\node[anchor=north,align=center] at ([yshift=-3mm]group c2r1.south) {GPT-5.4};
\node[anchor=north,align=center] at ([yshift=-3mm]group c3r1.south) {GPT-5.5};
\node[anchor=north,align=center] at ([yshift=-3mm]group c4r1.south) {Claude Opus~4.8};
\draw[regcontinue,solid,line width=1pt] ([xshift=-65.5mm,yshift=5mm]{$(group c1r1.north)!0.5!(group c4r1.north)$}) -- ++(5mm,0);
\draw[regcontinue,mark=*,mark size=1.5pt,mark options={fill=regcontinue}] plot coordinates {([xshift=-63mm,yshift=5mm]{$(group c1r1.north)!0.5!(group c4r1.north)$})};
\node[anchor=west] at ([xshift=-60mm,yshift=5mm]{$(group c1r1.north)!0.5!(group c4r1.north)$}) {Always continue};
\draw[regrestart,solid,line width=1pt] ([xshift=-25.5mm,yshift=5mm]{$(group c1r1.north)!0.5!(group c4r1.north)$}) -- ++(5mm,0);
\draw[regrestart,mark=square*,mark size=1.35pt,mark options={fill=regrestart}] plot coordinates {([xshift=-23mm,yshift=5mm]{$(group c1r1.north)!0.5!(group c4r1.north)$})};
\node[anchor=west] at ([xshift=-20mm,yshift=5mm]{$(group c1r1.north)!0.5!(group c4r1.north)$}) {Always restart};
\draw[regfixed,solid,line width=1pt] ([xshift=13.5mm,yshift=5mm]{$(group c1r1.north)!0.5!(group c4r1.north)$}) -- ++(5mm,0);
\draw[regfixed,mark=triangle,mark size=2.0pt,mark options={fill=white}] plot coordinates {([xshift=16mm,yshift=5mm]{$(group c1r1.north)!0.5!(group c4r1.north)$})};
\node[anchor=west] at ([xshift=19mm,yshift=5mm]{$(group c1r1.north)!0.5!(group c4r1.north)$}) {Best fixed};
\draw[regrde,solid,line width=1pt] ([xshift=44.5mm,yshift=5mm]{$(group c1r1.north)!0.5!(group c4r1.north)$}) -- ++(5mm,0);
\draw[regrde,mark=diamond*,mark size=1.9pt,mark options={fill=regrde}] plot coordinates {([xshift=47mm,yshift=5mm]{$(group c1r1.north)!0.5!(group c4r1.north)$})};
\node[anchor=west] at ([xshift=50mm,yshift=5mm]{$(group c1r1.north)!0.5!(group c4r1.north)$}) {R-DE};
\node[rotate=90,anchor=south] at ([xshift=-6mm]group c1r1.west) {Regret};
\node[anchor=north] at ([yshift=-7.5mm]{$(group c1r1.south)!0.5!(group c4r1.south)$}) {Inference budget};
\end{tikzpicture}
\endgroup
\caption{\textbf{Next-block regret across inference budgets.} Solved trials lost; lower is better. Budgets include the additional block.}
\label{fig:checkpoint-regret}
\end{figure}
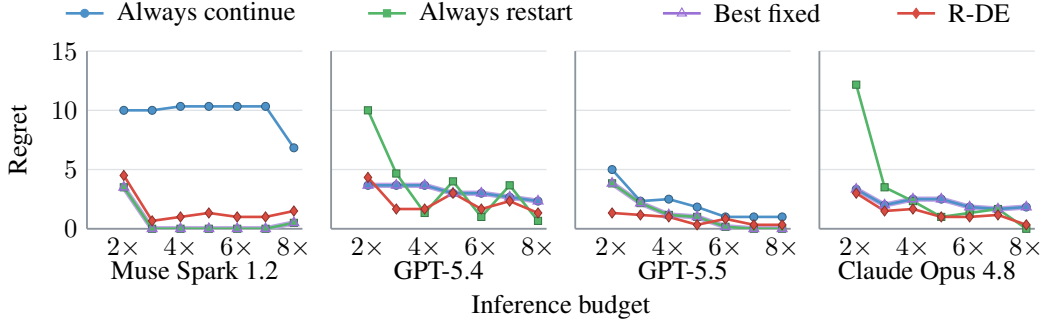

Across seven checkpoints, R-DE's mean measured next-block regret is 1.50, versus 1.79 for each model's retrospectively selected best fixed action, 4.30 for always continuing, and 2.13 for always restarting (Table~\ref{tab:checkpoint-regret}). R-DE modestly improves on best fixed for both GPT models and Opus, but not Muse.

R-DE's largest average gains over best fixed occur at $3\times$ and $4\times$ (Figure~\ref{fig:checkpoint-regret}; Appendix~\ref{app:decision-uncertainty}), avoiding much of the initial restart cost for Opus and GPT-5.4.

\section{Discussion and limitations}
\label{sec:robustness}
R-DE's aggregate forecasts show that conditional execution adds predictive information beyond short-budget success. SG remains competitive near first-block saturation, whereas Opus benefits from modeling completion over longer budgets. Gains vary across models and problems (Appendix~\ref{app:problem-level}). Muse's sketch-timing sensitivity and preference for reduced late discovery reveal limits of the assumptions (Appendix~\ref{app:robustness}). DE provides a predictive decomposition rather than an account of internal reasoning.

Main forecasts use problem-specific oracle sketches, but execution information also transfers across problems (Appendix~\ref{app:prior-sensitivity}). The 57-problem sample and three oracle trajectories per problem limit precision, particularly for individual execution curves and small policy differences.

\section{Conclusion}
We introduced a Discovery--Execution framework that models mathematical proof success as a convolution of strategy discovery and conditional execution. Measurements from short unaided attempts and Oracle-Execution trajectories predict aggregate scaling under held-out compute allocations without fitting the target outcomes. When execution is certain, the framework recovers geometric scaling, while incorporating compute-dependent execution improves beyond geometric extrapolation. The same measurements inform continue-or-restart decisions, yielding lower average regret than each model's best fixed action. Together, these results show that short-budget success alone does not determine the value of longer reasoning: measuring conditional execution provides additional information for predicting scaling and informing compute allocation.

\clearpage
\subsection*{AI use statement}

OpenAI Codex was used to propose and refine hypotheses, critique experimental design, assist with the formulation and analysis of the mathematical framework and arguments, implement and debug the experimental harness and analysis code, analyze and interpret results, survey related literature, and prepare the manuscript. Language models also serve as solver and grading instruments in the experiments, as described in the paper. The authors reviewed all AI-assisted work and take responsibility for the final text, claims, code, and artifacts.

\subsection*{Reproducibility statement}

We will publicly release the experimental harness, audit and framework-fitting code, and AOBench, including its reference proofs, strategy sketches, and load-bearing steps. Section~\ref{sec:experiments} describes the models, compute budgets, and evaluation protocol. Appendices~\ref{app:dataset-subtopics}, \ref{app:propositions}, \ref{app:joint-estimator}, and \ref{app:human-validation} provide dataset details, theoretical proofs, estimation procedures, and audit protocols, respectively.

\bibliographystyle{iclr2027_conference}
\bibliography{references}

\appendix
\section{Dataset and protocol details}
\label{app:dataset-subtopics}

\FloatBarrier
\subsection{Datasets}

AOBench was fixed before model evaluation and contains 35 non-geometry problems from competitions held in 2026. The sources are the China TST (12 problems), Iran TST (5), All-Russian MO (3), IMO (3), Korea FKMO (2), Romanian Master of Mathematics (2), Romania TST (2), Serbian MO (2), APMO (1), ELMO Shortlist (1), Serbia TST (1), and USAMO (1). Each record retains the original source and a complete checked reference solution. The problems were selected by the IMO medalist panel to be comparable in difficulty and size to the IMO-ProofBench Advanced dataset. No problem was removed based on model performance. Geometry was excluded before evaluation because the extensive algebraic calculations produced by models make human audit impractical.

\begin{figure}[H]
\centering
\definecolor{topicblue}{HTML}{356AA0}
\definecolor{topicorange}{HTML}{D97706}
\definecolor{topicpurple}{HTML}{7A5195}
\definecolor{topicgreen}{HTML}{3A8D6D}
\definecolor{topicred}{HTML}{C44E52}
\definecolor{topicviolet}{HTML}{8172B3}
\definecolor{topicgold}{HTML}{CCB974}
\definecolor{topiccyan}{HTML}{4C9DAE}

\newcommand{\topicentry}[2]{%
  \raggedright\scriptsize\sffamily
  \tikz[baseline=-0.5ex]{\filldraw[fill=#1,draw=black] (0,0) rectangle (1.2ex,1.2ex);}%
  \hspace{0.3ex}#2%
}

\begin{minipage}[t]{0.32\linewidth}
\centering
\begin{tikzpicture}[font=\normalfont\footnotesize\sffamily]
  \pie[sum=auto,radius=1.55,rotate=90,text=inside,hide label,change direction, color={topicblue,topicorange,topicpurple}] {5/Polynomial,3/Functional equation,1/Inequality}
\end{tikzpicture}
\vspace{1mm}

\begin{minipage}[t][27mm][t]{\linewidth}
\begin{tabular}{@{}p{0.46\linewidth}@{\hspace{0.04\linewidth}}p{0.46\linewidth}@{}} \topicentry{topicblue}{Polynomial} & \topicentry{topicorange}{Functional equation} \tabularnewline
\topicentry{topicpurple}{Inequality} &
\end{tabular}
\end{minipage}
\par\vspace{1mm}
\subcaption{Algebra ($n=9$)}
\end{minipage}
\hfill
\begin{minipage}[t]{0.32\linewidth}
\centering
\begin{tikzpicture}[font=\normalfont\footnotesize\sffamily]
  \pie[sum=auto,radius=1.55,rotate=90,text=inside,hide label, color={topicblue,topicorange,topicpurple,topicgreen,topicred,topicviolet,topicgold}] {3/Sequence,2/Diophantine equation,2/Divisibility, 2/Functional equation,2/Representation, 1/Number theoretic functions,1/Polynomial}
\end{tikzpicture}
\vspace{1mm}

\begin{minipage}[t][27mm][t]{\linewidth}
\begin{tabular}{@{}p{0.46\linewidth}@{\hspace{0.04\linewidth}}p{0.46\linewidth}@{}} \topicentry{topicblue}{Sequence} & \topicentry{topicgreen}{Functional equation} \tabularnewline
\topicentry{topicorange}{Diophantine equation} & \topicentry{topicpurple}{Divisibility} \tabularnewline
\topicentry{topicred}{Representation} & \topicentry{topicgold}{Polynomial} \tabularnewline
\topicentry{topicviolet}{Number theoretic functions} &
\end{tabular}
\end{minipage}
\par\vspace{1mm}
\subcaption{Number theory ($n=13$)}
\end{minipage}
\hfill
\begin{minipage}[t]{0.32\linewidth}
\centering
\begin{tikzpicture}[font=\normalfont\footnotesize\sffamily]
  \pie[sum=auto,radius=1.55,rotate=90,text=inside,hide label, color={topicblue,topicorange,topicpurple,topicgreen,topicred,topicviolet,topicgold,topiccyan}] {3/Extremal combinatorics,2/Game theory,2/Graph theory,2/Tiling, 1/Additive combinatorics,1/Combinatorial geometry, 1/Enum. combinatorics,1/Set combinatorics}
\end{tikzpicture}
\vspace{1mm}

\begin{minipage}[t][27mm][t]{\linewidth}
\begin{tabular}{@{}p{0.46\linewidth}@{\hspace{0.04\linewidth}}p{0.46\linewidth}@{}} \topicentry{topicblue}{Extremal combinatorics} & \topicentry{topicorange}{Game theory} \tabularnewline
\topicentry{topicpurple}{Graph theory} & \topicentry{topicred}{Additive combinatorics} \tabularnewline
\topicentry{topicgreen}{Tiling} & \topicentry{topicviolet}{Combinatorial geometry} \tabularnewline
\topicentry{topicgold}{Enum. combinatorics} & \topicentry{topiccyan}{Set combinatorics}
\end{tabular}
\end{minipage}
\par\vspace{1mm}
\subcaption{Combinatorics ($n=13$)}
\end{minipage}

\caption{AOBench subtopics. Slice labels give problem counts; geometry is excluded from both benchmark cohorts.}
\label{fig:dataset-subtopics}
\end{figure}
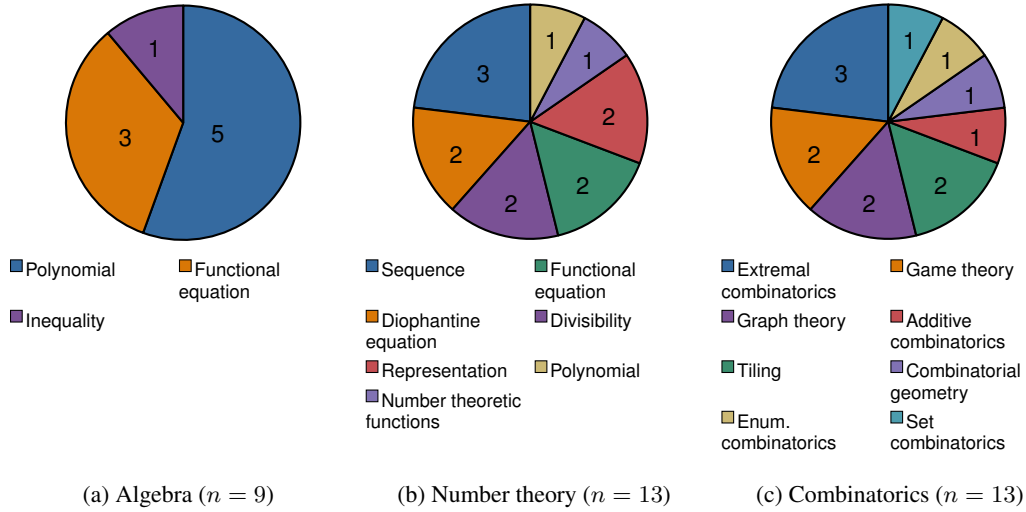

The replication set contains all 22 algebra, combinatorics, and number-theory problems in the Advanced split of IMO-ProofBench. Geometry was excluded by the same pre-specified rule. The resulting domain counts are 8 algebra, 8 combinatorics, and 6 number theory.

\FloatBarrier
\subsection{Harness and stopping protocol}
\label{app:harness}
\label{app:early-stopping}

The Python harness uses Claude Agent SDK version 0.1.56, with automatic context compaction configured at 900k tokens. Solvers execute code in a local scratch directory; WebSearch and WebFetch tools are blocked. Appendix~\ref{app:audit-example} gives a full worked example.

After a trajectory stops, later budget checkpoints retain its last complete proof. Cumulative success records whether any evaluated checkpoint has a passing proof; an incorrect terminal proof does not erase an earlier success. Among 171 trajectories per model and arm, self-converged runs with no passing checkpoint number 38, 22, 0, and 4 for unaided Muse, GPT-5.4, GPT-5.5, and Opus, respectively, and 27, 7, 1, and 0 for their oracle arms. 

\section{Proofs and derivations}
\label{app:propositions}

\subsection{Proof of Proposition 1}
\label{app:proposition1}

Since $\varepsilon$ is cumulative, saturated first-block execution, $\varepsilon(1)=1$, implies $\varepsilon(\ell)=1$ for every $\ell\geq1$. Substituting geometric discovery into Eq.~\ref{eq:discovery-execution-convolution} gives
\[
s(K)=\sum_{k=1}^{K}\alpha(1-\alpha)^{k-1}=1-(1-\alpha)^K.
\]
At one block, $q=\alpha\varepsilon(1)=\alpha$. For $N$ independent attempts with $NK=B$, the probability that at least one solves is therefore
\[
1-(1-s(K))^N=1-(1-q)^{NK}=1-(1-q)^B,
\]
which is independent of the allocation.
\hfill$\square$

\paragraph{Extension to flat, unsaturated execution.} More generally, if $\varepsilon(\ell)=c\in(0,1]$ for every $\ell\geq1$, then $K=1$, $N=B$ is an optimal allocation, and is uniquely optimal when $c<1$. This distinguishes saturated execution from execution that remains flat below one. Substituting geometric discovery and constant execution probability into Eq.~\ref{eq:discovery-execution-convolution} gives
\begin{align*}
s(K) &=c\sum_{k=1}^{K}p(k) \\
&=c\alpha\sum_{k=1}^{K}(1-\alpha)^{k-1} \\
&=c\left[1-(1-\alpha)^K\right].
\end{align*}
Let $N_K=\lfloor B/K\rfloor$. Consequently, the success probability is
\[
1-\left[1-c+c(1-\alpha)^K\right]^{N_K}.
\]
By convexity of $x\mapsto x^K$,
\begin{align*}
1-c+c(1-\alpha)^K &=(1-c)1^K+c(1-\alpha)^K \\
&\geq \left[(1-c)1+c(1-\alpha)\right]^K \\
&=\left[1-c+c(1-\alpha)\right]^K.
\end{align*}
Raising both sides to $N_K$ yields
\[
\left[1-c+c(1-\alpha)^K\right]^{N_K} \geq \left[1-c+c(1-\alpha)\right]^{K N_K}.
\]
Because $KN_K\leq B$ and the bracketed quantity lies in $[0,1]$,
\[
\left[1-c+c(1-\alpha)\right]^{K N_K} \geq \left[1-c+c(1-\alpha)\right]^B.
\]
For $0<c<1$ and $K>1$, the inequality is strict; subtracting both sides from one proves that $K=1$ uniquely maximizes success. If $c=1$, equality holds in the convexity step, and the allocation has success probability
\[
1-(1-\alpha)^{K\lfloor B/K\rfloor}.
\]
This equals the $K=1$ value $1-(1-\alpha)^B$ whenever $K\lfloor B/K\rfloor=B$, and is otherwise weakly smaller.
\hfill$\square$

\subsection{Proof of Proposition 2}
\label{app:proposition2}

Under geometric discovery, Eq.~\ref{eq:geometric-discovery-execution} gives
\[
s(K)=\sum_{k=1}^{K}\alpha(1-\alpha)^{k-1} \varepsilon(K-k+1).
\]
For fixed $K$, as $\alpha\to0$, $(1-\alpha)^{k-1}=1+O(\alpha)$, so
\begin{equation}
s(K)=\alpha\sum_{\ell=1}^{K}\varepsilon(\ell) +O(\alpha^2).
\label{eq:rare-discovery-single}
\end{equation}
Let $N_K=\lfloor B/K\rfloor$. Since $s(K)=O(\alpha)$,
\begin{align}
S_B(K) &=1-\bigl(1-s(K)\bigr)^{N_K} \notag\\
&=\alpha N_K\sum_{\ell=1}^{K}\varepsilon(\ell) +O(\alpha^2). \label{eq:rare-discovery-allocation}
\end{align}
Define
\[
\bar\varepsilon(K)=\frac{1}{K}\sum_{\ell=1}^{K}\varepsilon(\ell).
\]
By definition, $\varepsilon(K)$ is non-decreasing: completing a proof within $K$ blocks also constitutes completing it within $K+1$ blocks. Consequently,
\[
\bar\varepsilon(K+1)-\bar\varepsilon(K) =\frac{\varepsilon(K+1)-\bar\varepsilon(K)}{K+1}\geq0.
\]
Moreover, the assumed strict improvement implies $\bar\varepsilon(K)<\bar\varepsilon(B)$ for every $K<B$. Since $N_KK\leq B$, for every $K<B$,
\[
N_KK\bar\varepsilon(K)<B\bar\varepsilon(B).
\]
Hence the first-order coefficient in Eq.~\ref{eq:rare-discovery-allocation} is uniquely maximized by $K=B$, for which $N_B=1$.
\hfill$\square$

\section{Framework estimation and prediction}
\label{app:joint-estimator}

For problem $n$, let $u_n$ of $m_n$ independent one-block unaided attempts succeed. Let $c_n^{(j)}$ count oracle trajectories first solving in block $j$, with category $B+1$ denoting no success through block $B$. Write $r_n=\sum_{j=1}^{B+1}c_n^{(j)}$. Our experiments use $m_n=24$, $r_n=3$, and $B=8$. Success is cumulative: a passing proof at any earlier checkpoint counts as success thereafter. All fits below use these measurement runs, never the target unaided outcomes. We write $S_n(N,K)$ for the probability that at least one of $N$ independent $K$-block arms succeeds.

\FloatBarrier
\subsection{Simple Geometric (SG)}
\label{app:sg-estimator}

Under SG, each block is an independent opportunity with success probability $q_n$, so $S_n^{\rm SG}(N,K)=1-(1-q_n)^{NK}$. Rather than substitute $\widehat q_n=u_n/m_n$, we use the finite-sample pass@$NK$ estimator
\begin{equation}
\widehat S_n^{\rm SG}(N,K)=1-\frac{\binom{m_n-u_n}{NK}}{\binom{m_n}{NK}},\qquad NK\leq m_n.
\label{eq:app-sg-prediction}
\end{equation}
The ratio is the fraction of size-$NK$ subsets of observed attempts containing only failures; its numerator is zero when $m_n-u_n<NK$. For independent identically distributed short attempts, this estimates $1-(1-q_n)^{NK}$ without plug-in bias. SG uses no oracle observations and assigns equal success probability to allocations with the same total budget.

\FloatBarrier
\subsection{Discovery--Execution (DE)}
\label{app:de-estimator}

DE separates one-block success into discovery and execution, $q_n=\alpha_n\varepsilon_n(1)$. Define the empirical execution curve $\widehat\varepsilon_n(\ell)=r_n^{-1}\sum_{j=1}^{\ell}c_n^{(j)}$. We estimate discovery by dividing the short-attempt success rate by first-block oracle completion, subject to $0\leq\alpha_n\leq1$:
\begin{equation}
\widehat\alpha_n=
\begin{cases}
0, & \widehat q_n=\widehat\varepsilon_n(1)=0,\\
\widehat q_n/\widehat\varepsilon_n(1), & \widehat q_n\leq\widehat\varepsilon_n(1),\quad\widehat\varepsilon_n(1)>0,\\
1, & \widehat q_n>\widehat\varepsilon_n(1).
\end{cases}
\label{eq:app-de-cases}
\end{equation}
We denote the adjusted execution curve by $\varepsilon_n^\star$. In the first two cases, $\varepsilon_n^\star(\ell)=\widehat\varepsilon_n(\ell)$. When both first-block rates are zero, discovery is unidentified; choosing zero is an implementation convention and yields zero predicted unaided success.

In the third case, no $\alpha_n\in[0,1]$ matches the fresh success rate while retaining the empirical oracle curve. We set $\widehat\alpha_n=1$, so fresh and oracle first-block outcomes estimate the same execution probability. Pooling their counts gives
\begin{equation}
\varepsilon_n^\star(1)=\frac{u_n+c_n^{(1)}}{m_n+r_n}.
\label{eq:app-de-pooled-execution}
\end{equation}
Changing first-block completion also changes the fraction of trajectories available to solve later. We preserve the observed completion rate \emph{among trajectories unsolved after block one}. Of the $r_n-c_n^{(1)}$ such oracle trajectories, $\sum_{j=2}^{\ell}c_n^{(j)}$ solve by block $\ell$. We estimate the conditional completion probability as
\begin{equation}
\delta_n(\ell)=\frac{\sum_{j=2}^{\ell}c_n^{(j)}}{r_n-c_n^{(1)}}
=\frac{\widehat\varepsilon_n(\ell)-\widehat\varepsilon_n(1)}{1-\widehat\varepsilon_n(1)}.
\label{eq:app-de-conditional-tail}
\end{equation}
Applying this conditional probability to the probability remaining after block one gives the adjusted cumulative execution probability:
\begin{equation}
\varepsilon_n^\star(\ell)=\underbrace{\varepsilon_n^\star(1)}_{\text{solved in block one}}
+\underbrace{\bigl(1-\varepsilon_n^\star(1)\bigr)\delta_n(\ell)}_{\text{solved later, by block }\ell},\qquad 1\leq\ell\leq B.
\label{eq:app-de-adjusted-execution}
\end{equation}
This preserves the relative frequencies of later completion and noncompletion and keeps the curve nondecreasing. Predictions then use the fitted discovery--execution convolution:
\begin{equation}
\widehat s_n(K)=\sum_{j=1}^{K}\widehat\alpha_n(1-\widehat\alpha_n)^{j-1}\varepsilon_n^\star(K-j+1),\qquad
\widehat S_n^{\rm DE}(N,K)=1-\bigl(1-\widehat s_n(K)\bigr)^N.
\label{eq:app-de-prediction}
\end{equation}

\FloatBarrier
\subsection{Regularized Simple Geometric (R-SG)}
\label{app:rsg-estimator}

R-SG models each problem's success probability using a shared Beta prior. For problem $n$, the prior and likelihood of $u_n$ successes in $m_n$ short attempts are
\begin{equation}
q_n\sim\operatorname{Beta}(a,b),\qquad L_n(q_n)=\binom{m_n}{u_n}q_n^{u_n}(1-q_n)^{m_n-u_n}.
\label{eq:app-rsg-likelihood}
\end{equation}
For each LLM, we determine $(\widehat a,\widehat b)$ to maximize the product of expected likelihoods across all 57 problems. Writing $\operatorname{B}(\cdot,\cdot)$ for the beta function,
\begin{equation}
\begin{aligned}
(\widehat a,\widehat b)&=\underset{a,b>0}{\arg\max}\sum_{n=1}^{57}\log\mathbb E_{q_n\sim\operatorname{Beta}(a,b)}\!\left[L_n(q_n)\right]\\
&=\underset{a,b>0}{\arg\max}\sum_{n=1}^{57}\left[\log\operatorname{B}(a+u_n,b+m_n-u_n)-\log\operatorname{B}(a,b)\right].
\end{aligned}
\label{eq:app-rsg-prior}
\end{equation}
We optimize $\log a,\log b\in[-12,12]$ with SciPy.

With the fitted prior fixed, conjugacy between the Beta prior and binomial likelihood gives the posterior in closed form:
\begin{equation}
q_n\mid u_n,m_n\sim\operatorname{Beta}\!\left(\widehat a+u_n,\widehat b+m_n-u_n\right).
\label{eq:app-rsg-posterior}
\end{equation}
Averaging geometric success over this posterior gives the final estimator:
\begin{equation}
\begin{aligned}
\widehat S_n^{\rm R\text{-}SG}(N,K)&=1-\mathbb E\!\left[(1-q_n)^{NK}\mid u_n,m_n\right]\\
&=1-\frac{\operatorname{B}(\widehat a+u_n,\widehat b+m_n-u_n+NK)}{\operatorname{B}(\widehat a+u_n,\widehat b+m_n-u_n)}\\
&=1-\prod_{j=0}^{NK-1}\frac{\widehat b+m_n-u_n+j}{\widehat a+\widehat b+m_n+j}.
\end{aligned}
\label{eq:app-rsg-prediction}
\end{equation}
\FloatBarrier
\subsection{Regularized Discovery--Execution (R-DE)}
\label{app:rde-estimator}

R-DE uses the same empirical-Bayes approach as R-SG, but learns separate distributions for discovery and execution. Let $\pi_n^{(j)}$ be the probability that execution first completes in block $j$, with category $B+1$ denoting noncompletion within the budget. Then $\varepsilon_n(\ell)=\sum_{j=1}^{\ell}\pi_n^{(j)}$. We place independent priors on discovery probability and execution outcomes:
\begin{equation}
\alpha_n\sim\operatorname{Beta}(a,b),\qquad \boldsymbol\pi_n\sim\operatorname{Dirichlet}(\boldsymbol d).
\label{eq:app-rde-priors}
\end{equation}
The two types of measurement runs inform different parts of this model. A short unaided attempt succeeds only if discovery and execution both succeed within the first block, with probability $\alpha_n\pi_n^{(1)}$. Oracle runs supply the strategy and therefore observe execution directly: $c_n^{(j)}$ counts trajectories that first complete in block $j$. Combining these observations gives the likelihood, up to coefficients depending only on the data:
\begin{equation}
L_n(\alpha_n,\boldsymbol\pi_n)\propto
[\alpha_n\pi_n^{(1)}]^{u_n}
[1-\alpha_n\pi_n^{(1)}]^{m_n-u_n}
\prod_{j=1}^{B+1}[\pi_n^{(j)}]^{c_n^{(j)}}.
\label{eq:app-de-likelihood}
\end{equation}
As in R-SG, we fit the shared prior by maximizing the product of expected likelihoods across the 57 problems for each LLM. Here both the Beta and Dirichlet shapes are fitted jointly:
\begin{equation}
(\widehat a,\widehat b,\widehat{\boldsymbol d})
=\underset{a,b,\boldsymbol d}{\arg\max}\sum_{n=1}^{57}\log
\mathbb E_{\substack{\alpha_n\sim\operatorname{Beta}(a,b)\\
\boldsymbol\pi_n\sim\operatorname{Dirichlet}(\boldsymbol d)}}
\!\left[L_n(\alpha_n,\boldsymbol\pi_n)\right].
\label{eq:app-rde-prior-fit}
\end{equation}
For the main fit, the Dirichlet distribution is defined on the execution categories observed in the pooled oracle measurements. We fit positive shapes for those categories and set the probabilities of all other categories to zero. Zero entries in $\boldsymbol d$ are bookkeeping for these omitted coordinates, not parameters of a full-dimensional Dirichlet density; Dirichlet expressions below are understood on the retained coordinates. We optimize the retained log-shapes with SciPy using the same bounds as R-SG. The transfer experiment in Appendix~\ref{app:prior-sensitivity} instead retains all $B+1$ categories, allowing outcomes absent from the prior-fitting problems to occur on held-out problems.

With the fitted prior fixed, each problem's observations $\mathcal D_n=(u_n,m_n,\boldsymbol c_n)$ determine its posterior. The oracle update is immediate: execution follows $\operatorname{Dirichlet}(\widehat{\boldsymbol d}+\boldsymbol c_n)$ before incorporating the short attempts. The short-attempt update is less direct because a failure does not reveal whether discovery or execution failed.

To express this uncertainty, let $h_n$ count the failed short attempts in which discovery succeeded but execution did not finish. If this count were known, discovery would have $u_n+h_n$ successes and $m_n-u_n-h_n$ failures. First-block execution would have $u_n$ successes and $h_n$ failures in addition to the oracle observations. Conjugacy then gives independent conditional posteriors:
\begin{equation}
\begin{aligned}
\alpha_n\mid h_n,\mathcal D_n
&\sim\operatorname{Beta}\!\left(\widehat a+u_n+h_n,\widehat b+m_n-u_n-h_n\right),\\
\pi_n^{(1)}\mid h_n,\mathcal D_n
&\sim\operatorname{Beta}\!\left(\widehat d_1+c_n^{(1)}+u_n,\sum_{j=2}^{B+1}(\widehat d_j+c_n^{(j)})+h_n\right).
\end{aligned}
\label{eq:app-rde-conditional-posterior}
\end{equation}
Among outcomes unfinished after block one, the relative execution probabilities $(\pi_n^{(j)}/(1-\pi_n^{(1)}))_{j>1}$ retain the Dirichlet distribution with shapes $(\widehat d_j+c_n^{(j)})_{j>1}$, independently of the two Beta variables for fixed $h_n$. These proportions come from the oracle update because short attempts do not observe later completion. Since $h_n$ is unknown, the full posterior averages these conditional distributions over $h_n=0,\ldots,m_n-u_n$, weighted by each count's posterior probability given the observations.

Prediction now follows the same principle as R-SG: average the success formula over the posterior. For each possible $h_n$, we average DE success over discovery and execution uncertainty, then average those predictions over $h_n$:
\begin{equation}
\widehat S_n^{\rm R\text{-}DE}(N,K)
=\mathbb E_{h_n\mid\mathcal D_n}\!\left[
\mathbb E_{\alpha_n,\boldsymbol\pi_n\mid h_n,\mathcal D_n}\!\left[
1-\bigl(1-s_n(K;\alpha_n,\boldsymbol\pi_n)\bigr)^N
\right]\right].
\label{eq:app-rde-prediction}
\end{equation}
Here $s_n(K;\alpha_n,\boldsymbol\pi_n)$ is the single-arm DE success probability from Eq.~\ref{eq:geometric-discovery-execution}. The result is the posterior predictive probability that at least one of $N$ arms solves problem $n$ within $K$ blocks each. The inner expectation is evaluated exactly using closed-form Beta and Dirichlet moments; the outer expectation is a finite weighted sum over the possible failure counts.

\section{Additional prediction results}
\label{app:prediction-details}

\FloatBarrier
\Needspace{0.65\textheight}
\subsection{Predictions across compute allocations}

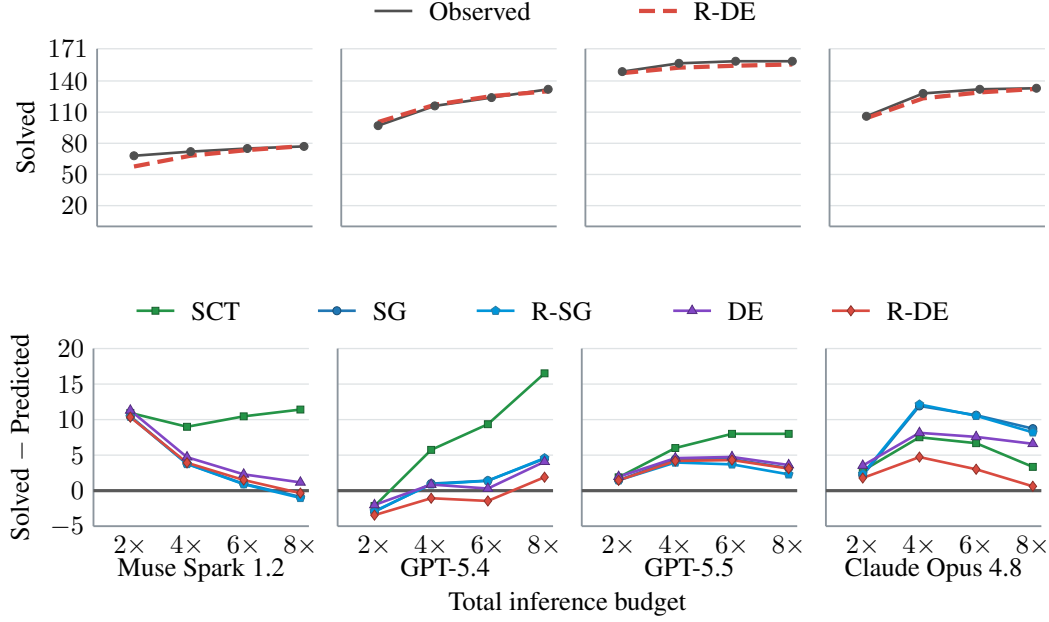
\begin{figure}[H]
\centering
\begingroup
\definecolor{n2observed}{HTML}{505050}
\definecolor{n2rde}{HTML}{D94B40}
\definecolor{n2axis}{HTML}{8C959D}
\definecolor{n2grid}{HTML}{D7DCDF}
\begin{tikzpicture}[font=\normalfont\normalsize,text=black]
\begin{groupplot}[group style={group size=4 by 1,horizontal sep=0.38cm},scale only axis,width=0.204\linewidth,height=2.35cm,xmin=0.7,xmax=8.3,ymin=0,ymax=171,xtick={2,4,6,8},xticklabels=\empty,ytick={20,50,80,110,140,171},tick label style={font=\normalfont\normalsize,text=black},label style={font=\normalfont\normalsize,text=black},axis line style={draw=n2axis,line width=0.7pt},axis x line*=bottom,axis y line*=left,tick style={draw=none},ymajorgrids=true,grid style={draw=n2grid,opacity=0.8,line width=0.5pt},clip=true]
\nextgroupplot[]
\addplot[draw=n2observed,solid,line width=1pt,mark=*,mark size=1.5pt,mark options={fill=n2observed,draw=n2observed,line width=0.25pt}] coordinates {(2,68.000000) (4,72.000000) (6,75.000000) (8,77.000000)};
\addplot[draw=n2rde,dash pattern=on 6pt off 3pt,line width=1.6pt,no marks] coordinates {(2,57.657077) (4,68.074023) (6,73.501733) (8,77.342466)};
\nextgroupplot[yticklabels=\empty]
\addplot[draw=n2observed,solid,line width=1pt,mark=*,mark size=1.5pt,mark options={fill=n2observed,draw=n2observed,line width=0.25pt}] coordinates {(2,97.000000) (4,116.000000) (6,124.000000) (8,132.000000)};
\addplot[draw=n2rde,dash pattern=on 6pt off 3pt,line width=1.6pt,no marks] coordinates {(2,100.445378) (4,117.065162) (6,125.447622) (8,130.103730)};
\nextgroupplot[yticklabels=\empty]
\addplot[draw=n2observed,solid,line width=1pt,mark=*,mark size=1.5pt,mark options={fill=n2observed,draw=n2observed,line width=0.25pt}] coordinates {(2,149.000000) (4,157.000000) (6,159.000000) (8,159.000000)};
\addplot[draw=n2rde,dash pattern=on 6pt off 3pt,line width=1.6pt,no marks] coordinates {(2,147.547136) (4,152.817832) (6,154.694011) (8,155.896154)};
\nextgroupplot[yticklabels=\empty]
\addplot[draw=n2observed,solid,line width=1pt,mark=*,mark size=1.5pt,mark options={fill=n2observed,draw=n2observed,line width=0.25pt}] coordinates {(2,106.000000) (4,128.000000) (6,132.000000) (8,133.000000)};
\addplot[draw=n2rde,dash pattern=on 6pt off 3pt,line width=1.6pt,no marks] coordinates {(2,104.219884) (4,123.271982) (6,128.989752) (8,132.388760)};
\end{groupplot}
\node[rotate=90,anchor=south] at ([xshift=-7mm]group c1r1.west) {Solved};
\draw[n2observed,solid,line width=1pt] ([xshift=-26mm,yshift=5mm]{$(group c1r1.north)!0.5!(group c4r1.north)$}) -- ++(5mm,0);
\node[anchor=west] at ([xshift=-20mm,yshift=5mm]{$(group c1r1.north)!0.5!(group c4r1.north)$}) {Observed};
\draw[n2rde,dash pattern=on 6pt off 3pt,line width=1.6pt] ([xshift=9mm,yshift=5mm]{$(group c1r1.north)!0.5!(group c4r1.north)$}) -- ++(5mm,0);
\node[anchor=west] at ([xshift=15mm,yshift=5mm]{$(group c1r1.north)!0.5!(group c4r1.north)$}) {R-DE};
\end{tikzpicture}
\endgroup
\par\vspace{6mm}
\begingroup
\definecolor{errorogt}{HTML}{249447}
\colorlet{errorrsg}{cyan!80!blue}
\definecolor{errorgeo}{HTML}{2077B4}
\definecolor{errorde}{HTML}{8246C5}
\definecolor{errorrde}{HTML}{D94B40}
\definecolor{errortick}{HTML}{44505A}
\definecolor{errorgroup}{HTML}{17212B}
\definecolor{erroraxis}{HTML}{8C959D}
\definecolor{errorgrid}{HTML}{D7DCDF}
\begin{tikzpicture}[font=\normalfont\normalsize,text=black]
\begin{groupplot}[group style={group size=4 by 1,horizontal sep=0.38cm},scale only axis,width=0.204\linewidth,height=2.35cm,xmin=0.7,xmax=8.3,ymin=-5,ymax=20,xtick={2,4,6,8},xticklabels={$2\times$,$4\times$,$6\times$,$8\times$},ytick={-5,0,5,10,15,20},tick label style={font=\normalfont\normalsize,text=black},xticklabel style={font=\normalfont\normalsize,text=black},label style={font=\normalfont\normalsize,text=black},axis line style={draw=erroraxis,line width=0.7pt},axis x line*=bottom,axis y line*=left,tick style={draw=none},ymajorgrids=true,grid style={draw=errorgrid,opacity=0.8,line width=0.5pt},clip=true]
\nextgroupplot[]
\addplot[black!65,solid,line width=1.2pt,no marks] coordinates {(0.7,0) (8.3,0)};
\addplot[draw=errorogt,solid,line width=1pt,mark=square*,mark size=1.3pt,mark options={solid,fill=errorogt,draw=errorogt!65!black,line width=0.25pt}] coordinates {(2,10.942708) (4,8.990162) (6,10.459491) (8,11.414931)};
\addplot[draw=errorgeo,solid,line width=1pt,mark=*,mark size=1.5pt,mark options={solid,fill=errorgeo,draw=errorgeo!65!black,line width=0.25pt}] coordinates {(2,10.467391) (4,3.800678) (6,0.964085) (8,-0.871576)};
\addplot[draw=errorrsg,solid,line width=1pt,mark=pentagon*,mark size=1.7pt,mark options={solid,fill=errorrsg,draw=errorrsg!65!black,line width=0.25pt}] coordinates {(2,10.417069) (4,3.756908) (6,0.906156) (8,-1.021226)};
\addplot[draw=errorde,solid,line width=1pt,mark=triangle*,mark size=2.1pt,mark options={solid,fill=errorde,draw=errorde!65!black,line width=0.25pt}] coordinates {(2,11.299447) (4,4.712186) (6,2.295047) (8,1.165381)};
\addplot[draw=errorrde,solid,line width=1pt,mark=diamond*,mark size=1.9pt,mark options={solid,fill=errorrde,draw=errorrde!65!black,line width=0.25pt}] coordinates {(2,10.342923) (4,3.925977) (6,1.498267) (8,-0.342466)};
\nextgroupplot[yticklabels=\empty]
\addplot[black!65,solid,line width=1.2pt,no marks] coordinates {(0.7,0) (8.3,0)};
\addplot[draw=errorogt,solid,line width=1pt,mark=square*,mark size=1.3pt,mark options={solid,fill=errorogt,draw=errorogt!65!black,line width=0.25pt}] coordinates {(2,-2.171875) (4,5.741319) (6,9.353009) (8,16.512731)};
\addplot[draw=errorgeo,solid,line width=1pt,mark=*,mark size=1.5pt,mark options={solid,fill=errorgeo,draw=errorgeo!65!black,line width=0.25pt}] coordinates {(2,-2.967391) (4,0.992095) (6,1.407642) (8,4.490461)};
\addplot[draw=errorrsg,solid,line width=1pt,mark=pentagon*,mark size=1.7pt,mark options={solid,fill=errorrsg,draw=errorrsg!65!black,line width=0.25pt}] coordinates {(2,-2.887362) (4,0.891998) (6,1.397821) (8,4.578501)};
\addplot[draw=errorde,solid,line width=1pt,mark=triangle*,mark size=2.1pt,mark options={solid,fill=errorde,draw=errorde!65!black,line width=0.25pt}] coordinates {(2,-2.010867) (4,0.860504) (6,0.299480) (8,4.073581)};
\addplot[draw=errorrde,solid,line width=1pt,mark=diamond*,mark size=1.9pt,mark options={solid,fill=errorrde,draw=errorrde!65!black,line width=0.25pt}] coordinates {(2,-3.445378) (4,-1.065162) (6,-1.447622) (8,1.896270)};
\nextgroupplot[yticklabels=\empty]
\addplot[black!65,solid,line width=1.2pt,no marks] coordinates {(0.7,0) (8.3,0)};
\addplot[draw=errorogt,solid,line width=1pt,mark=square*,mark size=1.3pt,mark options={solid,fill=errorogt,draw=errorogt!65!black,line width=0.25pt}] coordinates {(2,1.875000) (4,5.999421) (6,7.999421) (8,7.999421)};
\addplot[draw=errorgeo,solid,line width=1pt,mark=*,mark size=1.5pt,mark options={solid,fill=errorgeo,draw=errorgeo!65!black,line width=0.25pt}] coordinates {(2,1.467391) (4,4.434783) (6,4.430481) (8,3.182340)};
\addplot[draw=errorrsg,solid,line width=1pt,mark=pentagon*,mark size=1.7pt,mark options={solid,fill=errorrsg,draw=errorrsg!65!black,line width=0.25pt}] coordinates {(2,1.468622) (4,3.963021) (6,3.699899) (8,2.294904)};
\addplot[draw=errorde,solid,line width=1pt,mark=triangle*,mark size=2.1pt,mark options={solid,fill=errorde,draw=errorde!65!black,line width=0.25pt}] coordinates {(2,1.961741) (4,4.577583) (6,4.742158) (8,3.607806)};
\addplot[draw=errorrde,solid,line width=1pt,mark=diamond*,mark size=1.9pt,mark options={solid,fill=errorrde,draw=errorrde!65!black,line width=0.25pt}] coordinates {(2,1.452864) (4,4.182168) (6,4.305989) (8,3.103846)};
\nextgroupplot[yticklabels=\empty]
\addplot[black!65,solid,line width=1.2pt,no marks] coordinates {(0.7,0) (8.3,0)};
\addplot[draw=errorogt,solid,line width=1pt,mark=square*,mark size=1.3pt,mark options={solid,fill=errorogt,draw=errorogt!65!black,line width=0.25pt}] coordinates {(2,2.812500) (4,7.515046) (6,6.681134) (8,3.347801)};
\addplot[draw=errorgeo,solid,line width=1pt,mark=*,mark size=1.5pt,mark options={solid,fill=errorgeo,draw=errorgeo!65!black,line width=0.25pt}] coordinates {(2,2.163043) (4,11.922643) (6,10.620494) (8,8.731368)};
\addplot[draw=errorrsg,solid,line width=1pt,mark=pentagon*,mark size=1.7pt,mark options={solid,fill=errorrsg,draw=errorrsg!65!black,line width=0.25pt}] coordinates {(2,2.358885) (4,12.135214) (6,10.536020) (8,8.206544)};
\addplot[draw=errorde,solid,line width=1pt,mark=triangle*,mark size=2.1pt,mark options={solid,fill=errorde,draw=errorde!65!black,line width=0.25pt}] coordinates {(2,3.533179) (4,8.139127) (6,7.569328) (8,6.597201)};
\addplot[draw=errorrde,solid,line width=1pt,mark=diamond*,mark size=1.9pt,mark options={solid,fill=errorrde,draw=errorrde!65!black,line width=0.25pt}] coordinates {(2,1.780116) (4,4.728018) (6,3.010248) (8,0.611240)};
\end{groupplot}
\node[font=\normalfont\normalsize,anchor=north] at ([yshift=-3mm]group c1r1.south) {Muse Spark~1.2};
\node[font=\normalfont\normalsize,anchor=north] at ([yshift=-3mm]group c2r1.south) {GPT-5.4};
\node[font=\normalfont\normalsize,anchor=north] at ([yshift=-3mm]group c3r1.south) {GPT-5.5};
\node[font=\normalfont\normalsize,anchor=north] at ([yshift=-3mm]group c4r1.south) {Claude Opus~4.8};
\node[rotate=90,anchor=south,font=\normalfont\normalsize] at ([xshift=-7mm]group c1r1.west) {Solved $-$ Predicted};
\node[anchor=north,font=\normalfont\normalsize] at ([yshift=-7.5mm]{$(group c1r1.south)!0.5!(group c4r1.south)$}) {Total inference budget};
\draw[errorogt,solid,line width=1pt] ([xshift=-57mm,yshift=5mm]{$(group c1r1.north)!0.5!(group c4r1.north)$}) -- ++(5mm,0);
\draw[errorogt,mark=square*,mark size=1.3pt,mark options={fill=errorogt,draw=errorogt!65!black,line width=0.25pt}] plot coordinates {([xshift=-54.5mm,yshift=5mm]{$(group c1r1.north)!0.5!(group c4r1.north)$})};
\node[anchor=west,font=\normalfont\normalsize] at ([xshift=-51mm,yshift=5mm]{$(group c1r1.north)!0.5!(group c4r1.north)$}) {SCT};
\draw[errorgeo,solid,line width=1pt] ([xshift=-33mm,yshift=5mm]{$(group c1r1.north)!0.5!(group c4r1.north)$}) -- ++(5mm,0);
\draw[errorgeo,mark=*,mark size=1.5pt,mark options={fill=errorgeo,draw=errorgeo!65!black,line width=0.25pt}] plot coordinates {([xshift=-30.5mm,yshift=5mm]{$(group c1r1.north)!0.5!(group c4r1.north)$})};
\node[anchor=west,font=\normalfont\normalsize] at ([xshift=-27mm,yshift=5mm]{$(group c1r1.north)!0.5!(group c4r1.north)$}) {SG};
\draw[errorrsg,solid,line width=1pt] ([xshift=-12mm,yshift=5mm]{$(group c1r1.north)!0.5!(group c4r1.north)$}) -- ++(5mm,0);
\draw[errorrsg,mark=pentagon*,mark size=1.7pt,mark options={fill=errorrsg,draw=errorrsg!65!black,line width=0.25pt}] plot coordinates {([xshift=-9.5mm,yshift=5mm]{$(group c1r1.north)!0.5!(group c4r1.north)$})};
\node[anchor=west,font=\normalfont\normalsize] at ([xshift=-6mm,yshift=5mm]{$(group c1r1.north)!0.5!(group c4r1.north)$}) {R-SG};
\draw[errorde,solid,line width=1pt] ([xshift=14mm,yshift=5mm]{$(group c1r1.north)!0.5!(group c4r1.north)$}) -- ++(5mm,0);
\draw[errorde,mark=triangle*,mark size=2.1pt,mark options={fill=errorde,draw=errorde!65!black,line width=0.25pt}] plot coordinates {([xshift=16.5mm,yshift=5mm]{$(group c1r1.north)!0.5!(group c4r1.north)$})};
\node[anchor=west,font=\normalfont\normalsize] at ([xshift=20mm,yshift=5mm]{$(group c1r1.north)!0.5!(group c4r1.north)$}) {DE};
\draw[errorrde,solid,line width=1pt] ([xshift=35mm,yshift=5mm]{$(group c1r1.north)!0.5!(group c4r1.north)$}) -- ++(5mm,0);
\draw[errorrde,mark=diamond*,mark size=1.9pt,mark options={fill=errorrde,draw=errorrde!65!black,line width=0.25pt}] plot coordinates {([xshift=37.5mm,yshift=5mm]{$(group c1r1.north)!0.5!(group c4r1.north)$})};
\node[anchor=west,font=\normalfont\normalsize] at ([xshift=41mm,yshift=5mm]{$(group c1r1.north)!0.5!(group c4r1.north)$}) {R-DE};
\end{tikzpicture}
\endgroup
\caption{\textbf{Predicting two-arm success ($N=2$).} Top: observed successes and R-DE predictions. Bottom: observed minus predicted solved counts. Each curve represents 171 allocation trials across 57 problems; the horizontal axis gives total compute across both arms.}
\label{fig:allocation-transfer-bars}
\end{figure}

Figure~\ref{fig:allocation-transfer-bars} shows the Parallel-2 held-out curves underlying Table~\ref{tab:predictor-comparison}. Each trial succeeds if either of its two independent trajectories produces a passing proof within the per-arm budget. Predictions use the same Parallel-8 and Oracle-Execution measurement runs that fit Parallel-1. For Muse, R-DE predicts 77.3 final successes against 77 observed; the largest gap is at the first checkpoint, with 57.7 predicted against 68 observed.

Table~\ref{tab:n4-predictor-comparison} extends the comparison to $N=4$ arms of $K=2$ blocks each for Muse and GPT-5.5. Each model contributes three allocation trials per problem, and success requires at least one of the four arms to solve. All predictors retain their measurement-run fits. For GPT-5.5, R-SG has the lowest error; for Muse, all predictors substantially underestimate success, including R-DE, which predicts 78.5 final successes against 93 observed. The additional allocation therefore exposes a transfer failure despite Muse's close final prediction at $N=2$. Appendix~\ref{app:muse-n4-diagnostics} diagnoses this error.

\begin{table}[H]
\caption{Prediction RMSE in solved-trial counts for $N=4$, evaluated across both checkpoints of four arms with two blocks each. Each model has 57 problems and 171 allocation trials. Predictions use the unchanged intervention fit. Bold marks row minima.}
\label{tab:n4-predictor-comparison}
\centering\small
\begin{tabular}{lrrrrr}
\toprule Model & SG & R-SG & SCT & DE & R-DE \\
\midrule
Muse Spark~1.2 & 17.06 & 16.97 & 19.40 & 18.77 & \textbf{16.64} \\
GPT-5.5 & 1.85 & \textbf{1.14} & 1.71 & 2.27 & 1.57 \\
\bottomrule
\end{tabular}
\end{table}

\FloatBarrier
\Needspace{0.75\textheight}
\subsection{Dataset-separated predictions}

\begin{figure}[H]
\centering
\begin{subfigure}{\linewidth}
\centering
\begingroup
\definecolor{fitblue}{HTML}{356AA0}
\definecolor{fitorange}{HTML}{D97706}
\definecolor{fitpurple}{HTML}{7A5195}
\begin{tikzpicture}[font=\sffamily\small]
\begin{groupplot}[group style={group size=4 by 1,horizontal sep=0.38cm},scale only axis,width=0.204\linewidth,height=2.5cm,xmin=0.7,xmax=8.3,ymin=0,ymax=171,xtick={2,4,6,8},xticklabels={$2\times$,$4\times$,$6\times$,$8\times$},ytick={0,50,100,150},tick label style={font=\sffamily\small},axis x line*=bottom,axis y line*=left,tick style={draw=none},axis line style={gray},ymajorgrids,grid style={gray!20},clip=true]
\nextgroupplot[]
\addplot[draw=fitblue,solid,line width=1.5pt,opacity=0.78,mark=*,mark size=1.3pt] coordinates {(1,28.000000) (2,31.000000) (3,31.000000) (4,31.000000) (5,31.000000) (6,31.000000) (7,31.000000) (8,33.000000)};
\addplot[draw=fitblue,dashed,line width=1.5pt,opacity=0.78,mark=o,mark size=1.3pt] coordinates {(1,21.192728) (2,28.176351) (3,31.192122) (4,33.097309) (5,35.040782) (6,36.082208) (7,36.883113) (8,37.612077)};
\addplot[draw=fitorange,solid,line width=1.5pt,opacity=0.78,mark=*,mark size=1.3pt] coordinates {(1,28.000000) (2,32.000000) (3,33.000000) (4,33.000000) (5,33.000000) (6,33.000000) (7,33.000000) (8,35.000000)};
\addplot[draw=fitorange,dashed,line width=1.5pt,opacity=0.78,mark=o,mark size=1.3pt] coordinates {(1,24.863217) (2,28.131241) (3,30.469463) (4,32.517331) (5,33.648551) (6,34.471277) (7,35.123608) (8,36.016341)};
\addplot[draw=fitpurple,solid,line width=1.5pt,opacity=0.78,mark=*,mark size=1.3pt] coordinates {(1,56.000000) (2,63.000000) (3,64.000000) (4,64.000000) (5,64.000000) (6,64.000000) (7,64.000000) (8,68.000000)};
\addplot[draw=fitpurple,dashed,line width=1.5pt,opacity=0.78,mark=o,mark size=1.3pt] coordinates {(1,46.055945) (2,56.307593) (3,61.661585) (4,65.614640) (5,68.689334) (6,70.553485) (7,72.006720) (8,73.628419)};
\nextgroupplot[yticklabels=\empty]
\addplot[draw=fitblue,solid,line width=1.5pt,opacity=0.78,mark=*,mark size=1.3pt] coordinates {(1,45.000000) (2,58.000000) (3,63.000000) (4,64.000000) (5,64.000000) (6,65.000000) (7,68.000000) (8,69.000000)};
\addplot[draw=fitblue,dashed,line width=1.5pt,opacity=0.78,mark=o,mark size=1.3pt] coordinates {(1,42.012041) (2,54.510713) (3,61.313685) (4,65.421730) (5,67.909363) (6,69.999917) (7,71.509854) (8,72.723827)};
\addplot[draw=fitorange,solid,line width=1.5pt,opacity=0.78,mark=*,mark size=1.3pt] coordinates {(1,39.000000) (2,45.000000) (3,46.000000) (4,47.000000) (5,52.000000) (6,52.000000) (7,53.000000) (8,53.000000)};
\addplot[draw=fitorange,dashed,line width=1.5pt,opacity=0.78,mark=o,mark size=1.3pt] coordinates {(1,39.141507) (2,45.955306) (3,49.390812) (4,50.971015) (5,51.889246) (6,52.628547) (7,53.116148) (8,53.491523)};
\addplot[draw=fitpurple,solid,line width=1.5pt,opacity=0.78,mark=*,mark size=1.3pt] coordinates {(1,84.000000) (2,103.000000) (3,109.000000) (4,111.000000) (5,116.000000) (6,117.000000) (7,121.000000) (8,122.000000)};
\addplot[draw=fitpurple,dashed,line width=1.5pt,opacity=0.78,mark=o,mark size=1.3pt] coordinates {(1,81.153547) (2,100.466019) (3,110.704497) (4,116.392746) (5,119.798609) (6,122.628464) (7,124.626002) (8,126.215350)};
\nextgroupplot[yticklabels=\empty]
\addplot[draw=fitblue,solid,line width=1.5pt,opacity=0.78,mark=*,mark size=1.3pt] coordinates {(1,80.000000) (2,87.000000) (3,91.000000) (4,92.000000) (5,94.000000) (6,95.000000) (7,95.000000) (8,95.000000)};
\addplot[draw=fitblue,dashed,line width=1.5pt,opacity=0.78,mark=o,mark size=1.3pt] coordinates {(1,81.268849) (2,86.715113) (3,88.356960) (4,89.405917) (5,90.184622) (6,91.304734) (7,91.838431) (8,92.289605)};
\addplot[draw=fitorange,solid,line width=1.5pt,opacity=0.78,mark=*,mark size=1.3pt] coordinates {(1,58.000000) (2,60.000000) (3,62.000000) (4,63.000000) (5,63.000000) (6,63.000000) (7,63.000000) (8,63.000000)};
\addplot[draw=fitorange,dashed,line width=1.5pt,opacity=0.78,mark=o,mark size=1.3pt] coordinates {(1,56.470967) (2,60.287679) (3,61.033051) (4,61.424361) (5,61.664169) (6,62.029693) (7,62.149941) (8,62.236167)};
\addplot[draw=fitpurple,solid,line width=1.5pt,opacity=0.78,mark=*,mark size=1.3pt] coordinates {(1,138.000000) (2,147.000000) (3,153.000000) (4,155.000000) (5,157.000000) (6,158.000000) (7,158.000000) (8,158.000000)};
\addplot[draw=fitpurple,dashed,line width=1.5pt,opacity=0.78,mark=o,mark size=1.3pt] coordinates {(1,137.739816) (2,147.002792) (3,149.390011) (4,150.830279) (5,151.848791) (6,153.334427) (7,153.988371) (8,154.525772)};
\nextgroupplot[yticklabels=\empty]
\addplot[draw=fitblue,solid,line width=1.5pt,opacity=0.78,mark=*,mark size=1.3pt] coordinates {(1,43.000000) (2,60.000000) (3,65.000000) (4,67.000000) (5,67.000000) (6,68.000000) (7,69.000000) (8,69.000000)};
\addplot[draw=fitblue,dashed,line width=1.5pt,opacity=0.78,mark=o,mark size=1.3pt] coordinates {(1,45.391445) (2,61.513155) (3,66.862151) (4,69.805793) (5,73.094632) (6,73.856847) (7,75.179322) (8,76.255260)};
\addplot[draw=fitorange,solid,line width=1.5pt,opacity=0.78,mark=*,mark size=1.3pt] coordinates {(1,48.000000) (2,55.000000) (3,56.000000) (4,57.000000) (5,58.000000) (6,59.000000) (7,60.000000) (8,60.000000)};
\addplot[draw=fitorange,dashed,line width=1.5pt,opacity=0.78,mark=o,mark size=1.3pt] coordinates {(1,42.619380) (2,48.125826) (3,50.791986) (4,52.130053) (5,53.699145) (6,54.127206) (7,55.043186) (8,55.398278)};
\addplot[draw=fitpurple,solid,line width=1.5pt,opacity=0.78,mark=*,mark size=1.3pt] coordinates {(1,91.000000) (2,115.000000) (3,121.000000) (4,124.000000) (5,125.000000) (6,127.000000) (7,129.000000) (8,129.000000)};
\addplot[draw=fitpurple,dashed,line width=1.5pt,opacity=0.78,mark=o,mark size=1.3pt] coordinates {(1,88.010824) (2,109.638981) (3,117.654136) (4,121.935847) (5,126.793777) (6,127.984053) (7,130.222508) (8,131.653538)};
\end{groupplot}
\node[anchor=north,font=\sffamily\small] at ([yshift=-4mm]group c1r1.south) {Muse Spark~1.2};
\node[anchor=north,font=\sffamily\small] at ([yshift=-4mm]group c2r1.south) {GPT-5.4};
\node[anchor=north,font=\sffamily\small] at ([yshift=-4mm]group c3r1.south) {GPT-5.5};
\node[anchor=north,font=\sffamily\small] at ([yshift=-4mm]group c4r1.south) {Claude Opus~4.8};
\node[rotate=90,anchor=south] at ([xshift=-7mm]group c1r1.west) {Cumulative solved};
\node[anchor=north] at ([yshift=-9mm]{$(group c1r1.south)!0.5!(group c4r1.south)$}) {Total inference budget};
\draw[fitblue,solid,line width=1.5pt] ([xshift=-48mm,yshift=5mm]{$(group c1r1.north)!0.5!(group c4r1.north)$}) -- ++(5mm,0);
\node[anchor=west,draw=none,fill=none,inner sep=0pt] at ([xshift=-41mm,yshift=5mm]{$(group c1r1.north)!0.5!(group c4r1.north)$}) {AOBench};
\draw[fitorange,solid,line width=1.5pt] ([xshift=-18mm,yshift=5mm]{$(group c1r1.north)!0.5!(group c4r1.north)$}) -- ++(5mm,0);
\node[anchor=west,draw=none,fill=none,inner sep=0pt] at ([xshift=-11mm,yshift=5mm]{$(group c1r1.north)!0.5!(group c4r1.north)$}) {IMO-ProofBench};
\draw[fitpurple,solid,line width=1.5pt] ([xshift=27mm,yshift=5mm]{$(group c1r1.north)!0.5!(group c4r1.north)$}) -- ++(5mm,0);
\node[anchor=west,draw=none,fill=none,inner sep=0pt] at ([xshift=34mm,yshift=5mm]{$(group c1r1.north)!0.5!(group c4r1.north)$}) {Combined};
\end{tikzpicture}
\endgroup
\caption{$N=1$: one arm, eight blocks.}
\end{subfigure}
\par\medskip
\begin{subfigure}{\linewidth}
\centering
\begingroup
\definecolor{fitblue}{HTML}{356AA0}
\definecolor{fitorange}{HTML}{D97706}
\definecolor{fitpurple}{HTML}{7A5195}
\begin{tikzpicture}[font=\sffamily\small]
\begin{groupplot}[group style={group size=4 by 1,horizontal sep=0.38cm},scale only axis,width=0.204\linewidth,height=2.5cm,xmin=0.7,xmax=8.3,ymin=0,ymax=171,xtick={2,4,6,8},xticklabels={$2\times$,$4\times$,$6\times$,$8\times$},ytick={0,50,100,150},tick label style={font=\sffamily\small},axis x line*=bottom,axis y line*=left,tick style={draw=none},axis line style={gray},ymajorgrids,grid style={gray!20},clip=true]
\nextgroupplot[]
\addplot[draw=fitblue,solid,line width=1.5pt,opacity=0.78,mark=*,mark size=1.3pt] coordinates {(2,34.000000) (4,35.000000) (6,37.000000) (8,39.000000)};
\addplot[draw=fitblue,dashed,line width=1.5pt,opacity=0.78,mark=o,mark size=1.3pt] coordinates {(2,28.329020) (4,35.287461) (6,37.974129) (8,39.680049)};
\addplot[draw=fitorange,solid,line width=1.5pt,opacity=0.78,mark=*,mark size=1.3pt] coordinates {(2,34.000000) (4,37.000000) (6,38.000000) (8,38.000000)};
\addplot[draw=fitorange,dashed,line width=1.5pt,opacity=0.78,mark=o,mark size=1.3pt] coordinates {(2,29.328057) (4,32.786562) (6,35.527604) (8,37.662418)};
\addplot[draw=fitpurple,solid,line width=1.5pt,opacity=0.78,mark=*,mark size=1.3pt] coordinates {(2,68.000000) (4,72.000000) (6,75.000000) (8,77.000000)};
\addplot[draw=fitpurple,dashed,line width=1.5pt,opacity=0.78,mark=o,mark size=1.3pt] coordinates {(2,57.657077) (4,68.074023) (6,73.501733) (8,77.342466)};
\nextgroupplot[yticklabels=\empty]
\addplot[draw=fitblue,solid,line width=1.5pt,opacity=0.78,mark=*,mark size=1.3pt] coordinates {(2,52.000000) (4,64.000000) (6,70.000000) (8,77.000000)};
\addplot[draw=fitblue,dashed,line width=1.5pt,opacity=0.78,mark=o,mark size=1.3pt] coordinates {(2,54.062319) (4,65.425745) (6,71.503191) (8,75.089425)};
\addplot[draw=fitorange,solid,line width=1.5pt,opacity=0.78,mark=*,mark size=1.3pt] coordinates {(2,45.000000) (4,52.000000) (6,54.000000) (8,55.000000)};
\addplot[draw=fitorange,dashed,line width=1.5pt,opacity=0.78,mark=o,mark size=1.3pt] coordinates {(2,46.383060) (4,51.639417) (6,53.944432) (8,55.014305)};
\addplot[draw=fitpurple,solid,line width=1.5pt,opacity=0.78,mark=*,mark size=1.3pt] coordinates {(2,97.000000) (4,116.000000) (6,124.000000) (8,132.000000)};
\addplot[draw=fitpurple,dashed,line width=1.5pt,opacity=0.78,mark=o,mark size=1.3pt] coordinates {(2,100.445378) (4,117.065162) (6,125.447622) (8,130.103730)};
\nextgroupplot[yticklabels=\empty]
\addplot[draw=fitblue,solid,line width=1.5pt,opacity=0.78,mark=*,mark size=1.3pt] coordinates {(2,89.000000) (4,94.000000) (6,96.000000) (8,96.000000)};
\addplot[draw=fitblue,dashed,line width=1.5pt,opacity=0.78,mark=o,mark size=1.3pt] coordinates {(2,87.312654) (4,90.472835) (6,91.916530) (8,92.904545)};
\addplot[draw=fitorange,solid,line width=1.5pt,opacity=0.78,mark=*,mark size=1.3pt] coordinates {(2,60.000000) (4,63.000000) (6,63.000000) (8,63.000000)};
\addplot[draw=fitorange,dashed,line width=1.5pt,opacity=0.78,mark=o,mark size=1.3pt] coordinates {(2,60.234482) (4,62.344997) (6,62.777480) (8,62.991609)};
\addplot[draw=fitpurple,solid,line width=1.5pt,opacity=0.78,mark=*,mark size=1.3pt] coordinates {(2,149.000000) (4,157.000000) (6,159.000000) (8,159.000000)};
\addplot[draw=fitpurple,dashed,line width=1.5pt,opacity=0.78,mark=o,mark size=1.3pt] coordinates {(2,147.547136) (4,152.817832) (6,154.694011) (8,155.896154)};
\nextgroupplot[yticklabels=\empty]
\addplot[draw=fitblue,solid,line width=1.5pt,opacity=0.78,mark=*,mark size=1.3pt] coordinates {(2,56.000000) (4,72.000000) (6,75.000000) (8,75.000000)};
\addplot[draw=fitblue,dashed,line width=1.5pt,opacity=0.78,mark=o,mark size=1.3pt] coordinates {(2,55.713894) (4,69.948559) (6,73.899149) (8,76.498403)};
\addplot[draw=fitorange,solid,line width=1.5pt,opacity=0.78,mark=*,mark size=1.3pt] coordinates {(2,50.000000) (4,56.000000) (6,57.000000) (8,58.000000)};
\addplot[draw=fitorange,dashed,line width=1.5pt,opacity=0.78,mark=o,mark size=1.3pt] coordinates {(2,48.505990) (4,53.323424) (6,55.090603) (8,55.890356)};
\addplot[draw=fitpurple,solid,line width=1.5pt,opacity=0.78,mark=*,mark size=1.3pt] coordinates {(2,106.000000) (4,128.000000) (6,132.000000) (8,133.000000)};
\addplot[draw=fitpurple,dashed,line width=1.5pt,opacity=0.78,mark=o,mark size=1.3pt] coordinates {(2,104.219884) (4,123.271982) (6,128.989752) (8,132.388760)};
\end{groupplot}
\node[anchor=north,font=\sffamily\small] at ([yshift=-4mm]group c1r1.south) {Muse Spark~1.2};
\node[anchor=north,font=\sffamily\small] at ([yshift=-4mm]group c2r1.south) {GPT-5.4};
\node[anchor=north,font=\sffamily\small] at ([yshift=-4mm]group c3r1.south) {GPT-5.5};
\node[anchor=north,font=\sffamily\small] at ([yshift=-4mm]group c4r1.south) {Claude Opus~4.8};
\node[rotate=90,anchor=south] at ([xshift=-7mm]group c1r1.west) {Cumulative solved};
\node[anchor=north] at ([yshift=-9mm]{$(group c1r1.south)!0.5!(group c4r1.south)$}) {Total inference budget};
\draw[fitblue,solid,line width=1.5pt] ([xshift=-48mm,yshift=5mm]{$(group c1r1.north)!0.5!(group c4r1.north)$}) -- ++(5mm,0);
\node[anchor=west,draw=none,fill=none,inner sep=0pt] at ([xshift=-41mm,yshift=5mm]{$(group c1r1.north)!0.5!(group c4r1.north)$}) {AOBench};
\draw[fitorange,solid,line width=1.5pt] ([xshift=-18mm,yshift=5mm]{$(group c1r1.north)!0.5!(group c4r1.north)$}) -- ++(5mm,0);
\node[anchor=west,draw=none,fill=none,inner sep=0pt] at ([xshift=-11mm,yshift=5mm]{$(group c1r1.north)!0.5!(group c4r1.north)$}) {IMO-ProofBench};
\draw[fitpurple,solid,line width=1.5pt] ([xshift=27mm,yshift=5mm]{$(group c1r1.north)!0.5!(group c4r1.north)$}) -- ++(5mm,0);
\node[anchor=west,draw=none,fill=none,inner sep=0pt] at ([xshift=34mm,yshift=5mm]{$(group c1r1.north)!0.5!(group c4r1.north)$}) {Combined};
\end{tikzpicture}
\endgroup
\caption{$N=2$: two arms, four blocks each.}
\end{subfigure}
\caption{\textbf{Dataset-separated scaling.} Solid lines show observed successes; dashed lines show R-DE predictions. Colors identify each dataset and their combined total.}
\label{fig:dataset-scaling}
\end{figure}
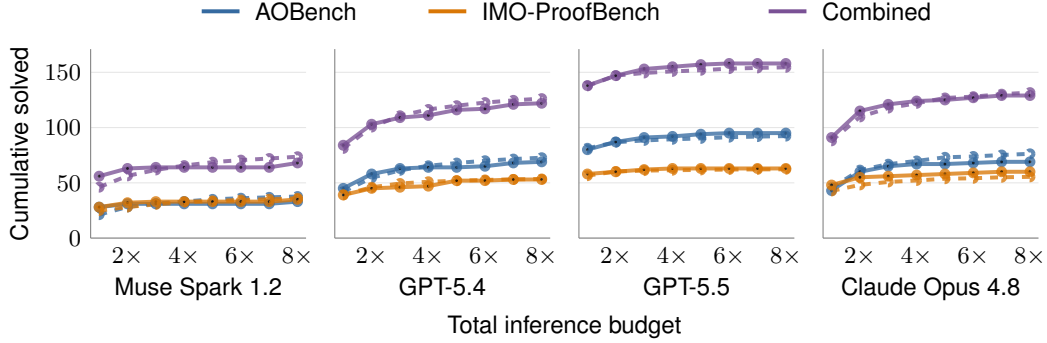
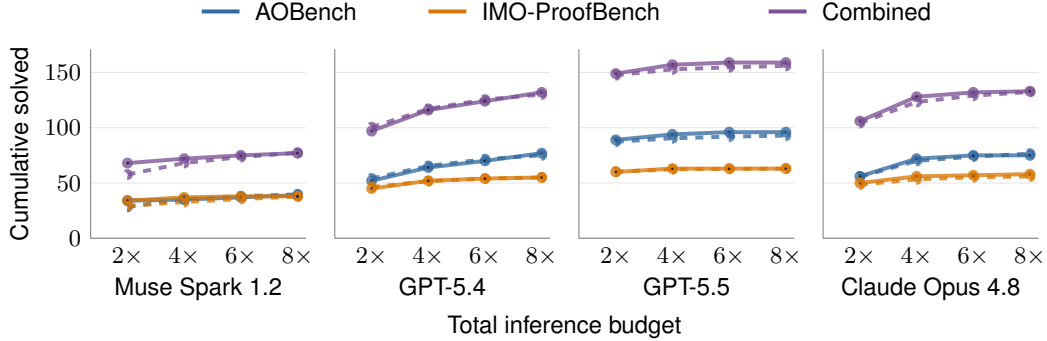

Figure~\ref{fig:dataset-scaling} separates the $N=1$ and $N=2$ results into AOBench and IMO-ProofBench. Each panel uses the same fitted parameters as the pooled evaluation and sums predictions and observed successes only over the indicated dataset. AOBench contributes 105 trials and IMO-ProofBench 66.

\FloatBarrier
\subsection{Problem-level prediction}
\label{app:problem-level}

Table~\ref{tab:problem-level} compares predictions with the six individual trajectories from Parallel-2 per problem at blocks $1$--$4$, predicting each trajectory with $N=1$. For each problem and checkpoint, we compare the predicted solved count with the observed count out of six. We square these errors, average over all $57\times4$ problem--checkpoint pairs, and take the square root. Fits use the same Parallel-8 and Oracle-Execution measurements as the main predictions.

\begin{table}[H]
\caption{Problem-level RMSE in solved-trajectory counts, using the six individual Parallel-2 trajectories per problem over blocks 1--4. Bold marks row minima.}
\label{tab:problem-level}
\centering\small
\begin{tabular}{lrrrrr}
\toprule Model & SG & R-SG & SCT & DE & R-DE \\
\midrule
Muse Spark~1.2 & 1.27 & 1.23 & 1.26 & 1.19 & \textbf{1.09} \\
GPT-5.4 & 1.22 & 1.21 & 1.51 & 1.20 & \textbf{1.16} \\
GPT-5.5 & 0.71 & \textbf{0.69} & 1.10 & 0.74 & 0.71 \\
Claude Opus~4.8 & 0.93 & 0.92 & 1.31 & 0.92 & \textbf{0.86} \\
\midrule
Average & 1.06 & 1.04 & 1.31 & 1.03 & \textbf{0.97} \\
\bottomrule
\end{tabular}
\end{table}

R-DE has the lowest problem-level RMSE for Muse, GPT-5.4, and Opus, while R-SG performs best for GPT-5.5. Problem-level improvements are smaller than aggregate improvements: for Opus, R-DE reduces RMSE from SG's 0.93 to 0.86 solved trajectories.

\FloatBarrier
\subsection{Prediction accuracy across allocations}
\label{app:prediction-uncertainty}

We compare the fitted predictors across the two target allocations using
\begin{equation}
\operatorname{RMSE}_{\mathrm{combined}}=\sqrt{\frac{\operatorname{RMSE}_{N=1}^{2}+\operatorname{RMSE}_{N=2}^{2}}{2}}.
\label{eq:combined-prediction-rmse}
\end{equation}
Each allocation has 171 trials across 57 problems. Allocation-specific MSE averages over eight checkpoints for $N=1$ and four for $N=2$, so Eq.~\ref{eq:combined-prediction-rmse} gives the allocations equal weight. This exploratory comparison follows the allocation-specific analysis in Table~\ref{tab:predictor-comparison}.

We hold the 57 problems and all fitted predictions fixed and perform 100,000 paired target-trial bootstrap replicates (random seed 1234). Within each problem, we sample three allocation trials with replacement, retaining each trial's complete cumulative success history. An $N=1$ trial is one eight-block trajectory; an $N=2$ trial is a recorded pair evaluated through four blocks per arm, successful if either arm solves. Every predictor is evaluated against the same resampled outcomes. Resampling preserves any shared-trajectory dependence across allocations, while separate target trials are resampled independently.

\begin{table}[H]
\caption{\textbf{Combined held-out prediction accuracy across $N=1$ and $N=2$.} Equal-allocation RMSE in solved-trial counts. Reductions are baseline minus R-DE, so positive values favor R-DE; brackets give 95\% paired target-bootstrap percentile intervals.}
\label{tab:combined-prediction-bootstrap}
\centering\small\setlength{\tabcolsep}{4pt}
\begin{tabular}{@{}lrrrll@{}}
\toprule
& \multicolumn{3}{c}{Combined RMSE} & \multicolumn{2}{c}{RMSE reduction relative to baseline} \\
\cmidrule(lr){2-4}\cmidrule(l){5-6}
Model & SG & R-SG & R-DE & SG minus R-DE & R-SG minus R-DE \\
\midrule
Muse & 7.06 & 7.08 & 5.93 & $1.13\;[-0.05,1.80]$ & $1.15\;[-0.06,1.84]$ \\
GPT-5.4 & 3.44 & 3.45 & 3.17 & $0.27\;[-0.85,1.26]$ & $0.28\;[-0.84,1.27]$ \\
GPT-5.5 & 3.05 & 2.52 & 3.56 & $-0.51\;[-0.77,0.18]$ & $-1.04\;[-1.27,-0.02]$ \\
Opus & 8.22 & 8.20 & 2.91 & $5.31\;[2.29,5.95]$ & $5.29\;[2.29,5.92]$ \\
\bottomrule
\end{tabular}
\end{table}

For Opus, R-DE lowers combined RMSE by 5.31 relative to SG and 5.29 relative to R-SG, with both percentile intervals above zero (Table~\ref{tab:combined-prediction-bootstrap}). The corresponding comparisons for Muse and GPT-5.4 include zero, while GPT-5.5 favors R-SG over R-DE. 

Pointwise 95\% percentile intervals quantify target-sampling uncertainty with the benchmark and fitted predictors fixed, without multiplicity adjustment. With three trials per problem and allocation, unanimous outcomes have zero bootstrap variance. Appendix~\ref{app:prior-sensitivity} separately assesses sensitivity to prior-fitting problems.

\FloatBarrier
\subsection{Continue-or-restart evaluation}
\label{app:decision-uncertainty}

At each checkpoint $k=1,\ldots,7$, the continuation count $C_n$ is the number of the problem's three Parallel-1 trajectories that first achieve a passing proof at $k+1$. The restart count $R_n$ estimates additional successes among trajectories still unsolved at $k$, using the six individual trajectories from Parallel-2 for that problem. These restart trajectories are separate from the continuation trajectories and are not used to fit the predictors. R-DE selects an action using its fitted expected advantage; regret compares the selected count with $\max\{C_n,R_n\}$, sums across problems, and averages across checkpoints, as in Eq.~\ref{eq:regret}.

These are descriptive regret comparisons; the additional supplement reports uncertainty analyses.

\section{Robustness and transfer}
\label{app:robustness}

\FloatBarrier
\subsection{Assumption 1: Execution independent of acquisition time}
\label{app:oracle-timing}

We compare no sketch, a sketch supplied at the start, and a sketch supplied after an unaided prefix. For each model, we select problem--seed trials whose recorded $3\times$ proof in the no-sketch arm scores below 5/7. No sketch and late sketch continue the same native prefix session for one additional block; start sketch uses the corresponding oracle $1\times$ checkpoint. 

\begin{table}[H]
\caption{AOBench sketch-timing comparison: solved/selected trajectories (score $\geq5/7$), with $\Delta\%=100(\mathrm{late}-\mathrm{start})/\mathrm{start}$. }
\label{tab:late-oracle}
\centering\small
\begin{tabular}{lrrrr}
\toprule Model & No sketch & Start sketch & Late sketch & $\Delta\%$ \\
\midrule
Muse Spark~1.2 & 1/71 & 34/71 & 17/71 & $-50.0\%$ \\
GPT-5.4 & 8/45 & 30/45 & 29/45 & $-3.3\%$ \\
GPT-5.5 & 3/12 & 11/12 & 11/12 & $0.0\%$ \\
Claude Opus~4.8 & 3/48 & 20/48 & 20/48 & $0.0\%$ \\
\bottomrule
\end{tabular}
\end{table}

The GPT start- and late-sketch counts differ by at most one solved trajectory, while Opus solves 20 trajectories in either sketch condition versus 3 without a sketch. Muse instead solves 34 trajectories with a sketch at the start and 17 with a late sketch, showing that its conditional completion is more sensitive to this change in context.

\FloatBarrier
\subsection{Assumption 2: Execution across viable strategies}
\label{app:alternate-sketches}

To assess whether the execution measurement depends on which reference strategy is supplied, we replace the original oracle sketch with a sketch of a human-checked AI-generated alternative proof on 38 problems: 24 from AOBench and 14 from IMO-ProofBench. Both sketches contain at most 25 words and use the same prompt format. Each condition has three standalone $1\times$ runs per problem for Muse, GPT-5.5, and Opus, giving 114 trajectories per model and condition. Both conditions are graded by GPT-5.6 Sol against their respective reference solutions; success means any valid proof scoring at least 5/7.

\begin{table}[H]
\caption{Solved $1\times$ trajectories on 38 matched problems (24 AOBench, 14 IMO-ProofBench), with three seeds each. $\Delta\%=100(\mathrm{alternate}-\mathrm{original})/\mathrm{original}$.}
\label{tab:alternate-sketch}
\centering\small
\begin{tabular}{lrrr}
\toprule Model & Original sketch & Alternate sketch & $\Delta\%$ \\
\midrule
Muse Spark~1.2 & 58/114 & 66/114 & $+13.8\%$ \\
GPT-5.5 & 109/114 & 111/114 & $+1.8\%$ \\
Claude Opus~4.8 & 88/114 & 82/114 & $-6.8\%$ \\
\bottomrule
\end{tabular}
\end{table}

Out of 114 trajectories per condition, Muse solves 58 with the original sketch and 66 with the alternate sketch; GPT-5.5 solves 109 and 111, respectively, and Opus solves 88 and 82.

\FloatBarrier
\subsection{Assumption 3: Geometric discovery}
\label{app:memoryless-acquisition}

We test sensitivity to reduced late discovery by replacing geometric discovery with
\begin{equation}
p_n^{(\gamma)}(k)=\alpha_n\bigl((1-\alpha_n)\gamma\bigr)^{k-1},\qquad 0\leq\gamma\leq1.
\label{eq:memoryless-decay}
\end{equation}
The case $\gamma=1$ recovers the original framework; smaller values reduce late discovery. We hold the measurement-fitted R-DE posterior fixed and fit one $\gamma$ per model to first-solve times from the 171 Parallel-1 trajectories, jointly integrating each problem's three trajectories over its shared parameters. We use 16,384 posterior draws per problem, with seeds 1234--1237 for Muse, GPT-5.4, GPT-5.5, and Opus, respectively.

\begin{table}[H]
\caption{Geometric-discovery sensitivity on all 57 problems. $\Delta\log L$ is the improvement over $\gamma=1$; the final column is the largest absolute change in predicted solved trajectories across the eight checkpoints.}
\label{tab:memoryless-acquisition}
\centering\small
\begin{tabular}{lrrrr}
\toprule Model & Trajectories & $\widehat\gamma$ & $\Delta\log L$ & Max. count change \\
\midrule
Muse Spark~1.2 & 171 & 0.692 & 7.58 & 12.48 \\
GPT-5.4 & 171 & 1.000 & 0.00 & 0.00 \\
GPT-5.5 & 171 & 0.972 & 0.19 & 0.79 \\
Claude Opus~4.8 & 171 & 0.974 & 0.12 & 0.86 \\
\bottomrule
\end{tabular}
\end{table}

Within this family, the GPT and Opus fits remain close to geometric discovery: $\widehat\gamma$ lies between 0.972 and 1, changing aggregate predictions by fewer than one solved trajectory. Muse instead favors reduced late discovery ($\widehat\gamma=0.692$), with a maximum change of 12.48 solved trajectories.

\FloatBarrier
\subsection{Sketch-content control}
\label{app:sketch-content}

To distinguish useful strategic content from simply adding text, we assign each problem the next problem's sketch in a sorted within-domain list, with cyclic wraparound. This preserves the sketch multiset and prompt wrapper. We compare no sketch, shuffled sketch, and the correct oracle sketch on all 35 AOBench problems, with three standalone $1\times$ runs per condition.

\begin{table}[H]
\caption{Sketch-content control on all 35 AOBench problems and three seeds. The last three columns count solved standalone $1\times$ trajectories (score $\geq5/7$); seeds are counted individually.}
\label{tab:placebo}
\centering\small
\begin{tabular}{lrrrr}
\toprule Model & Trajectories & No sketch & Shuffled sketch & Oracle sketch \\
\midrule
Muse Spark~1.2 & 105 & 24 & 13 & 48 \\
GPT-5.4 & 105 & 41 & 38 & 81 \\
GPT-5.5 & 105 & 83 & 77 & 104 \\
Claude Opus~4.8 & 105 & 45 & 41 & 73 \\
\bottomrule
\end{tabular}
\end{table}

Correct oracle sketches increase solved-trajectory counts for all four models; shuffled sketches do not improve over no sketch. 

\FloatBarrier
\subsection{First-block consistency}
\label{app:first-block-consistency}

We compare each problem's 24 one-block Parallel-8 attempts with the first-block outcomes of its three eight-block Parallel-1 trajectories. Across all 57 problems, these give 1,368 Parallel-8 outcomes and 171 Parallel-1 first-block outcomes per model. Table~\ref{tab:first-block-consistency} reports the percentage scoring at least 5/7 in each condition and their difference in percentage points. Every problem contributes the same number of trials within each condition, so these percentages also equal the averages of the problem-level success rates.

\begin{table}[H]
\caption{First-block success across all 57 problems: 1,368 Parallel-8 attempts and 171 Parallel-1 first-block outcomes per model. Differences are Parallel-1 minus Parallel-8, in percentage points (pp), computed before rounding.}
\label{tab:first-block-consistency}
\centering\small
\begin{tabular}{lrrr}
\toprule Model & Parallel-8 (\%) & Parallel-1, block 1 (\%) & Difference (pp) \\
\midrule
Muse Spark~1.2 & 27.0 & 32.7 & $+5.8$ \\
GPT-5.4 & 47.3 & 49.1 & $+1.8$ \\
GPT-5.5 & 80.6 & 80.7 & $+0.1$ \\
Claude Opus~4.8 & 51.6 & 53.2 & $+1.6$ \\
\bottomrule
\end{tabular}
\end{table}

Muse has the largest observed difference, with Parallel-1 first-block success 5.8 percentage points above Parallel-8 success. The other models differ by 0.1--1.8 percentage points.

\FloatBarrier
\subsection{Execution ablations}
\label{app:execution-ablations}

We ablate ordinary DE using the same measurement runs and held-out $N=1$ trajectories. \emph{Flat execution} retains each problem's fitted discovery rate but sets every execution checkpoint to its fitted first-block value. \emph{Shared execution} pools the model's 171 oracle trajectories into one empirical curve $\bar\varepsilon(k)$ and sets each problem's discovery rate to $\min\{1,\widehat q_n/\bar\varepsilon(1)\}$. The shared curve remains fixed across problems; target outcomes are used only for evaluation.

\begin{table}[H]
\caption{Execution ablations: $N=1$ RMSE in solved-trajectory counts across eight checkpoints, with 171 held-out trajectories per model. Lower is better; bold marks row minima.}
\label{tab:execution-ablations}
\centering\small
\begin{tabular}{lrrr}
\toprule Model & DE & Flat execution & Shared execution \\
\midrule
Muse Spark~1.2 & 6.54 & \textbf{5.22} & 17.17 \\
GPT-5.4 & \textbf{4.25} & 4.35 & 8.41 \\
GPT-5.5 & \textbf{3.04} & 5.31 & 5.67 \\
Claude Opus~4.8 & \textbf{3.68} & 19.47 & 21.20 \\
\bottomrule
\end{tabular}
\end{table}

Flattening execution substantially worsens Opus's predictions but improves Muse's, so measured execution gains do not transfer uniformly. Shared execution performs worse for every model. This ablation also constrains first-block fit: for Opus, 25 of 57 short success rates exceed $\bar\varepsilon(1)$. Its error therefore reflects the common execution level as well as the loss of problem-specific temporal variation.

\FloatBarrier
\subsection{Muse's four-arm prediction error}
\label{app:muse-n4-diagnostics}

R-DE predicts 78.5 final successes for Muse on Parallel-4, compared with 93 observed. Each trial has four arms of two blocks each. We diagnose this error with three post-hoc checks, changing one component at a time while keeping the shared priors fitted from the measurement runs fixed.

\textbf{First-block success.} Replace each problem's 24 one-block Parallel-8 outcomes with the first-block outcomes of its 12 Parallel-4 arms, keeping its oracle observations unchanged.

\textbf{Flat execution.} Hold the fitted execution curve at its first-block value, removing execution gains after block 1 while retaining the fitted discovery distribution.

\textbf{Discovery decay.} Apply $\widehat\gamma=0.692$, fitted to the eight-block Parallel-1 trajectories (Appendix~\ref{app:memoryless-acquisition}), which multiplies the geometric model's probability of discovery in block 2 by 0.692.

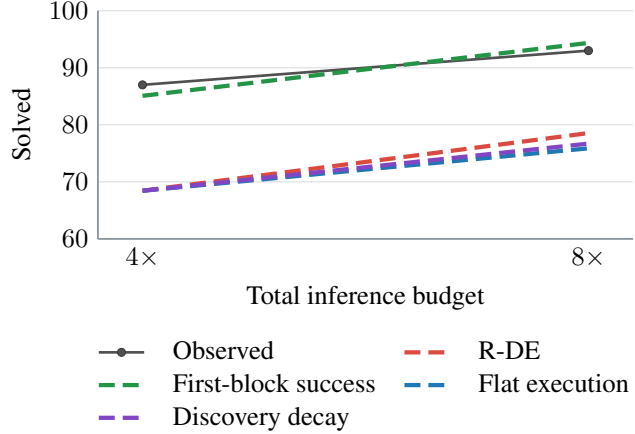
\begin{figure}[htbp]
\centering
\begingroup
\definecolor{plotunaided}{HTML}{505050}
\definecolor{errorrde}{HTML}{D94B40}
\definecolor{errorogt}{HTML}{249447}
\definecolor{errorgeo}{HTML}{2077B4}
\definecolor{errorde}{HTML}{8246C5}
\definecolor{erroraxis}{HTML}{8C959D}
\definecolor{errorgrid}{HTML}{D7DCDF}
\begin{tikzpicture}[font=\normalfont\normalsize,text=black]
\begin{axis}[width=0.62\linewidth,height=4.6cm,xmin=3.6,xmax=8.4,ymin=60,ymax=100,xtick={4,8},xticklabels={$4\times$,$8\times$},ytick={60,70,80,90,100},xlabel={Total inference budget},ylabel={Solved},tick label style={font=\normalfont\normalsize,text=black},label style={font=\normalfont\normalsize,text=black},axis x line*=bottom,axis y line*=left,axis line style={draw=erroraxis,line width=0.7pt},tick style={draw=none},ymajorgrids=true,grid style={draw=errorgrid,opacity=0.8,line width=0.5pt},legend style={at={(0.5,-0.4)},anchor=north,draw=none,font=\normalfont\normalsize,cells={anchor=west},column sep=8pt},legend columns=2]
\addplot[draw=plotunaided,solid,line width=1pt,mark=*,mark size=1.5pt,mark options={solid,fill=plotunaided,draw=plotunaided!65!black,line width=0.25pt}] coordinates {(4,87.000000) (8,93.000000)};
\addlegendentry{Observed}
\addplot[draw=errorrde,dash pattern=on 6pt off 3pt,line width=1.6pt,no marks] coordinates {(4,68.436321) (8,78.545287)};
\addlegendentry{R-DE}
\addplot[draw=errorogt,dash pattern=on 6pt off 3pt,line width=1.6pt,no marks] coordinates {(4,85.074847) (8,94.353027)};
\addlegendentry{First-block success}
\addplot[draw=errorgeo,dash pattern=on 6pt off 3pt,line width=1.6pt,no marks] coordinates {(4,68.436321) (8,75.875833)};
\addlegendentry{Flat execution}
\addplot[draw=errorde,dash pattern=on 6pt off 3pt,line width=1.6pt,no marks] coordinates {(4,68.436321) (8,76.699716)};
\addlegendentry{Discovery decay}
\end{axis}
\end{tikzpicture}
\endgroup
\caption{\textbf{Muse Parallel-4 observed and predicted successes.} Solved counts out of 171 trials for R-DE and the three post-hoc checks. Total budgets of $4\times$ and $8\times$ give one and two blocks per arm, respectively.}
\label{fig:muse-n4-diagnostics}
\end{figure}

Replacing first-block outcomes raises the final prediction to 94.4 and reduces RMSE from 16.6 to 1.7. Flat execution and discovery decay lower the final prediction to 75.9 and 76.7, respectively, worsening the error (Figure~\ref{fig:muse-n4-diagnostics}; Table~\ref{tab:muse-n4-diagnostics}).

\begin{table}[htbp]
\centering\small
\begin{tabular}{lrr}
\toprule R-DE condition & RMSE & Final predicted / observed \\
\midrule
R-DE & 16.6 & 78.5 / 93 \\
First-block success & 1.7 & 94.4 / 93 \\
Flat execution & 17.9 & 75.9 / 93 \\
Discovery decay & 17.5 & 76.7 / 93 \\
\bottomrule
\end{tabular}
\caption{Muse $N=4$ diagnostics. RMSE uses both checkpoints in Figure~\ref{fig:muse-n4-diagnostics}; final counts are at two blocks per arm, out of 171 trials.}
\label{tab:muse-n4-diagnostics}
\end{table}

\FloatBarrier

For Muse, the Parallel-8 attempts underestimate first-block success in the Parallel-4 arms: 27.0\% versus 34.8\%. Accounting for this difference removes most of the prediction error; the execution and discovery changes do not. The Parallel-1 trajectories show a similar first-block difference (32.7\% versus 27.0\%; Appendix~\ref{app:first-block-consistency}). These results point to a failure of the first-block measurement to transfer across allocations. Replacing first-block outcomes and fitting $\gamma$ use target information, so these are diagnostics rather than valid forecasts. Our system prompt states each attempt's total budget, which may affect Muse's behavior within the first block. We did not isolate this effect.

\FloatBarrier
\subsection{Transfer across problems}
\label{app:prior-sensitivity}
\label{app:no-test-oracle}

We split the 57 problems into 38 for prior fitting and 19 for evaluation, repeating this 50 times with the same splits for all four models (random seed 1234). Each split evaluates aggregate $N=1$ predictions at blocks $1$--$4$ on the six individual Parallel-2 trajectories per test problem, giving 114 test trajectories. Table~\ref{tab:prior-transfer} compares shared priors fitted to all 57 problems with priors fitted only to the 38 training problems, then tests prediction without oracle observations from the evaluation problems. Target outcomes are never used for fitting.

Without test oracle measurements, all counts $\boldsymbol c_n$ in Appendix~\ref{app:joint-estimator}, including noncompletion, are zero; predictions still average over the joint discovery--execution posterior. R-SG learns its Beta prior from the training problems' short runs.

\begin{table}[H]
\caption{\textbf{Transfer across problems.} Mean $\pm$ SD of solved-count RMSE over 50 splits, each using the six individual Parallel-2 trajectories for each of 19 test problems (114 trajectories), evaluated at blocks $1$--$4$ with $N=1$. All conditions except the first R-DE column fit priors on the 38 training problems.}
\label{tab:prior-transfer}
\centering\small
\begin{tabular}{lrrrr}
\toprule
& \multicolumn{3}{c}{R-DE} & R-SG \\
\cmidrule(lr){2-4}
Model & All 57 + oracle & Train 38 + oracle & Train 38, no oracle & Train 38 \\
\midrule
Muse Spark~1.2 & $4.57\pm1.90$ & $4.60\pm1.90$ & $4.62\pm2.08$ & $5.05\pm2.13$ \\
GPT-5.4 & $4.19\pm2.15$ & $4.16\pm2.05$ & $4.08\pm1.92$ & $4.12\pm2.12$ \\
GPT-5.5 & $3.78\pm1.83$ & $3.54\pm1.66$ & $3.56\pm1.53$ & $2.95\pm1.41$ \\
Claude Opus~4.8 & $3.62\pm1.47$ & $3.40\pm1.46$ & $5.04\pm2.67$ & $6.24\pm2.09$ \\
\bottomrule
\end{tabular}
\end{table}

Restricting prior fitting to 38 problems changes mean R-DE test RMSE by less than 0.25 solved trajectories for every model. Without test oracle measurements, R-DE improves on R-SG for Muse and Opus, is close for GPT-5.4, and performs worse for GPT-5.5. Opus benefits further from its own oracle measurements, reducing RMSE from 5.04 to 3.40. Execution information can therefore transfer to problems without supplied sketches. Standard deviations describe variation across overlapping splits, not uncertainty from independent replications.

\Needspace{6\baselineskip}
\section{Audit protocol and human validation}
\label{app:human-validation}
\begin{figure}[H]
\centering
\begin{tikzpicture}[x=1.25cm,y=0.9cm,font=\small]
\definecolor{auditblue}{HTML}{356AA0}
\foreach \row/\col/\pct/\txt in { 0/0/85.7/85.7,0/1/7.1/7.1,0/2/0/0,0/3/7.1/7.1, 1/0/0/0,1/1/16.7/16.7,1/2/50/50.0,1/3/33.3/33.3, 2/0/0/0,2/1/0/0,2/2/33.3/33.3,2/3/66.7/66.7, 3/0/0/0,3/1/0/0,3/2/0/0,3/3/100/100.0}{
  \filldraw[fill=auditblue!\pct!white,draw=gray!30] (\col,-\row) rectangle ++(1,-1);
  \ifdim\pct pt>60pt
    \node[text=white] at (\col+0.5,-\row-0.5) {\txt\%};
  \else
    \node at (\col+0.5,-\row-0.5) {\txt\%};
  \fi
}
\foreach \col/\score in {0/0,1/5,2/6,3/7}
  \node at (\col+0.5,0.3) {\score};
\foreach \row/\score/\total in {0/0/14,1/5/12,2/6/12,3/7/12}
  \node[anchor=east] at (-0.1,-\row-0.5) {\score\ ($n=\total$)};
\node at (2,0.85) {Human score};
\node[rotate=90] at (-1.65,-2) {Automated score};
\end{tikzpicture}
\caption{Automated versus human proof scores for all 50 completed audits. Percentages are normalized within each row; $n$ gives the number of proofs in that row.}
\label{fig:human-score-agreement}
\end{figure}

The proof judge receives the problem, submitted proof, and reference solution, without the solver identity or condition. Alternate proofs are allowed; the reference is not a required route. Scores $\geq5$ count as solved: minor local defects are allowed, but load-bearing gaps and missing cases fail. A separate tool-free state judge checks a frozen three-step outline, marking a step present only when its mechanism and role are explicit. These labels are diagnostic, not inputs to discovery estimation. 

An IMO medalist panel reviewed 50 anonymized proofs in disjoint sets. Each set contains seven automated score-0 proofs and six each with scores 5, 6, and 7; score-0 proofs are also stratified by reference-step presence. Auditors see the problem, reference solution, three-step outline, and submitted proof, but not solver identities or automated scores. We measure human--automated agreement on this stratified sample.

Solved status agrees on 48/50 proofs, giving 100.0\% precision, 94.7\% recall, and 97.3\% F1 against human labels. Thirty proof scores agree exactly; in every disagreement, the automated judge assigns the lower score (Figure~\ref{fig:human-score-agreement}). Reference-step presence agrees on 140/150 labels; these diagnostic labels are not used to fit the frameworks.

\FloatBarrier
\section{A worked example}
\label{app:audit-example}

We reproduce a GPT-5.5 Oracle-Execution trajectory on Romania TST 2026, Problem 7 (seed 3), item T21 in the human audit. The full reference proof, submitted proof, and audit judgments appear below with typesetting changes only.

\paragraph{Problem.} Do there exist a positive integer $N$ and an infinite sequence $a_1,a_2,a_3,\dots$ of positive integers such that
\[
a_{n+1}=\frac{a_1a_2\cdots a_n}{a_1+a_2+\cdots+a_n}\quad\text{for all }n\ge N\,?
\]

\paragraph{Supplied sketch.} Set $S_n=\sum_{i\le n}a_i$; derive $S_{n+1}\mid S_n^3$ and stabilize prime supports. For $p$, persistent $v_p(a_{n+1})\le v_p(S_n)$ forces unbounded $2$-divisibility; once reversed, the inequality persists. Combine stabilized valuations against growth.

\paragraph{Reference outline (audit only).}
\begin{enumerate}
\item Split the accumulated sum and the newest term by their gcd into coprime parts to constrain the next sum's prime support.
\item Rule out the persistent-equality branch by iterating its linear valuation recursion and examining divisibility constraints on a fixed exponent
\item Induct forward a strict valuation inequality between the new term and the accumulated quantity, uniformly over the stabilized prime set.
\end{enumerate}

\paragraph{Reference proof (audit only).}
The answer is \textbf{No}.

For $n \ge 1$, let $S_n = a_1 + a_2 + \dots + a_n$ and $P_n = a_1 a_2 \cdots a_n$.

Suppose there exist such $N$ and $(a_n)_{n \ge 1}$ satisfying $a_{n+1} = \frac{P_n}{S_n}$, i.e., $a_{n+1} S_n = P_n$ for all $n \ge N$.

Then for all $n \ge N$,

\[
a_{n+2} S_{n+1} = P_{n+1} = a_{n+1} P_n = a_{n+1}^2 S_n. \tag{1}
\]

Fix $n \ge N$. Let $g = \gcd(S_n, a_{n+1})$, so $S_n = gx$ and $a_{n+1} = gy$ for positive integers $x, y$ with $\gcd(x, y) = 1$. Then (1) gives $a_{n+2} g(x + y) = g^3 x y^2$, so $a_{n+2}(x + y) = g^2 x y^2$, implying $x + y \mid g^2$. Hence $S_{n+1} = g(x + y) \mid g^3 \mid S_n^3$.

So for all $n \ge N$, $S_{n+1} \mid S_n^3$. In particular, every prime dividing $S_{n+1}$ also divides $S_n$. Letting $A_n$ denote the set of primes dividing $S_n$, we get $A_n \supseteq A_{n+1}$ for all $n \ge N$. Since $A_N$ is finite, there exists $M \ge N$ such that $A_M = A_{M+1} = \cdots =: \mathcal{P}$.

Fix $p \in \mathcal{P}$. Applying $\nu_p$ to (1):

\[
\nu_p(a_{n+2}) + \nu_p(S_{n+1}) = 2\nu_p(a_{n+1}) + \nu_p(S_n). \tag{2}
\]

\textbf{Lemma.} There exists $n_0 \ge M$ such that $\nu_p(a_{n_0 + k + 1}) > \nu_p(S_{n_0 + k})$ for all integers $k \ge 0$.

\emph{Proof.} First we find $n_0 \ge M$ with $\nu_p(a_{n_0+1}) > \nu_p(S_{n_0})$. Assume for contradiction that $\nu_p(a_{n+1}) \le \nu_p(S_n)$ for all $n \ge M$. If $\nu_p(a_{t+1}) < \nu_p(S_t)$ for some $t \ge M$, then $\nu_p(S_{t+1}) = \nu_p(a_{t+1})$, and (2) at $n = t$ gives $\nu_p(a_{t+2}) = \nu_p(a_{t+1}) + \nu_p(S_t) > \nu_p(S_{t+1})$, contradicting the assumption. So $\nu_p(a_{n+1}) = \nu_p(S_n)$ for all $n \ge M$. Then (2) becomes $2\nu_p(S_{n+1}) = 3\nu_p(S_n)$, giving $2^l \nu_p(S_{M+l}) = 3^l \nu_p(S_M)$ for all $l \ge 0$. Hence $2^l \mid \nu_p(S_M)$ for all $l$, which is impossible since $\nu_p(S_M)$ is a fixed positive integer. So such $n_0$ exists.

Now by induction: if $\nu_p(a_{n_0+k+1}) > \nu_p(S_{n_0+k})$, then $\nu_p(S_{n_0+k+1}) = \nu_p(S_{n_0+k})$, and (2) gives $\nu_p(a_{n_0+k+2}) = 2\nu_p(a_{n_0+k+1}) > 2\nu_p(S_{n_0+k}) = 2\nu_p(S_{n_0+k+1}) > \nu_p(S_{n_0+k+1})$, completing the induction. $\blacksquare$

Applying the lemma for each $p \in \mathcal{P}$ (a finite set), choose $M' \ge M$ such that $\nu_p(a_{n+1}) > \nu_p(S_n)$ for all $n \ge M'$ and all $p \in \mathcal{P}$. Then $\nu_p(S_{n+1}) = \nu_p(S_n)$ for all such $n$ and $p$. Since $\mathcal{P}$ is the set of all prime divisors of $S_n$ (and of $S_{n+1}$), the numbers $S_n$ and $S_{n+1}$ have the same prime factorization, so $S_n = S_{n+1}$, meaning $a_{n+1} = 0$. This contradicts $(a_n)$ being a sequence of positive integers.

Therefore, no such $N$ and $(a_n)_{n \ge 1}$ exist.

\paragraph{Model-generated proof.}

The critique found no genuine gap. Here is the complete standalone solution.

\textbf{Final Solution}

No such sequence exists.

Assume, for contradiction, that there exist a positive integer \(N\) and an infinite sequence of positive integers \(a_1,a_2,a_3,\dots\) such that

\[
a_{n+1}=\frac{a_1a_2\cdots a_n}{a_1+a_2+\cdots+a_n} \tag{1}
\]

for every \(n\ge N\).

Let \(S_n=a_1+a_2+\cdots+a_n.\) For \(m\ge 0\), define \(X_m=S_{N+m},\qquad Y_m=a_{N+m+1}.\) Then

\[
X_{m+1}=S_{N+m+1}=S_{N+m}+a_{N+m+1}=X_m+Y_m. \tag{2}
\]

Applying (1) with \(n=N+m\), we get \(Y_m=a_{N+m+1} =\frac{a_1a_2\cdots a_{N+m}}{S_{N+m}} =\frac{a_1a_2\cdots a_{N+m}}{X_m}.\) Thus

\[
a_1a_2\cdots a_{N+m}=X_mY_m. \tag{3}
\]

Applying (1) with \(n=N+m+1\), and using (3), gives \(Y_{m+1} =a_{N+m+2} =\frac{a_1a_2\cdots a_{N+m+1}}{S_{N+m+1}} =\frac{X_mY_m^2}{X_{m+1}}.\) Hence

\[
X_{m+1}Y_{m+1}=X_mY_m^2. \tag{4}
\]

In particular,

\[
X_{m+1}\mid X_mY_m^2. \tag{5}
\]

Since \(X_{m+1}=X_m+Y_m\), we have \(Y_m\equiv -X_m \pmod {X_{m+1}}.\) Therefore \(X_mY_m^2\equiv X_m(-X_m)^2=X_m^3\pmod {X_{m+1}}.\) Because \(X_{m+1}\mid X_mY_m^2\), this implies

\[
X_{m+1}\mid X_m^3. \tag{6}
\]

Let \(\mathcal P\) be the set of prime divisors of \(X_0Y_0\). This set is finite. We claim that every prime divisor of every \(X_mY_m\) belongs to \(\mathcal P\).

The claim is true for \(m=0\) by definition. Suppose it is true for some \(m\). If a prime \(p\) divides \(X_{m+1}\), then by (5), \(p\mid X_mY_m^2.\) Since \(p\) is prime, \(p\mid X_m\) or \(p\mid Y_m\), so \(p\mid X_mY_m\). By the induction hypothesis, \(p\in\mathcal P\).

If a prime \(p\) divides \(Y_{m+1}\), then by (4), \(p\mid X_mY_m^2.\) Again \(p\mid X_m\) or \(p\mid Y_m\), so \(p\mid X_mY_m\), hence \(p\in\mathcal P\). Therefore every prime divisor of \(X_{m+1}Y_{m+1}\) belongs to \(\mathcal P\). By induction, every prime divisor of every \(X_mY_m\) belongs to \(\mathcal P\).

Now fix a prime \(p\in\mathcal P\). Define \(e_m=v_p(X_m),\qquad f_m=v_p(Y_m).\) Taking \(p\)-adic valuations in (4), we obtain \(e_{m+1}+f_{m+1}=e_m+2f_m,\) or equivalently,

\[
f_{m+1}=e_m+2f_m-e_{m+1}. \tag{7}
\]

We prove that \(e_m\) is eventually constant.

First suppose \(f_m>e_m\). Write \(X_m=p^{e_m}u,\qquad Y_m=p^{f_m}v,\) where \(p\nmid u\) and \(p\nmid v\). Then \(X_{m+1}=X_m+Y_m =p^{e_m}\left(u+p^{f_m-e_m}v\right).\) Since \(f_m-e_m>0\), the integer \(p^{f_m-e_m}v\) is divisible by \(p\), while \(u\) is not divisible by \(p\). Hence \(p\nmid u+p^{f_m-e_m}v,\) so \(e_{m+1}=e_m.\) Using (7), \(f_{m+1}=e_m+2f_m-e_{m+1}=2f_m.\) Thus \(f_{m+1}=2f_m>e_m=e_{m+1}.\) Therefore, once \(f_m>e_m\), the inequality \(f_t>e_t\) holds for every later index \(t\), and \(e_t\) is constant from that point onward.

Next suppose \(f_m<e_m\). Write \(X_m=p^{e_m}u,\qquad Y_m=p^{f_m}v,\) where \(p\nmid u\) and \(p\nmid v\). Then \(X_{m+1}=X_m+Y_m =p^{f_m}\left(p^{e_m-f_m}u+v\right).\) Since \(e_m-f_m>0\), the integer \(p^{e_m-f_m}u\) is divisible by \(p\), while \(v\) is not divisible by \(p\). Hence \(p\nmid p^{e_m-f_m}u+v,\) so \(e_{m+1}=f_m.\) Using (7), \(f_{m+1}=e_m+2f_m-e_{m+1}=e_m+f_m.\) Since \(e_m>f_m\ge 0\), \(f_{m+1}=e_m+f_m>f_m=e_{m+1}.\) Thus after one step we enter the previous case, so \(e_t\) is constant for all sufficiently large \(t\).

Now suppose \(e_m=f_m=0\). Then (7) gives \(f_{m+1}=-e_{m+1}.\) Both \(e_{m+1}\) and \(f_{m+1}\) are nonnegative integers, so \(e_{m+1}=f_{m+1}=0.\) Thus once \(e_m=f_m=0\), the same equality persists forever, and \(e_t\) is constant from that point onward.

It remains to consider the case \(e_m=f_m=E>0.\) Since both \(X_m\) and \(Y_m\) are divisible by \(p^E\), their sum \(X_{m+1}\) is divisible by \(p^E\). Hence \(e_{m+1}=E+h\) for some integer \(h\ge 0\). From (7),

\[
f_{m+1}=E+2E-(E+h)=2E-h. \tag{8}
\]

If \(f_{m+1}\ne e_{m+1}\), then at index \(m+1\) one of the two strict cases applies, and \(e_t\) is eventually constant.

Therefore, the only possible way for \(e_t\) not to become eventually constant is for equality \(e_t=f_t>0\) to persist forever from some point onward. Suppose this happens from an index \(m_0\). Write \(e_{m_0+r}=f_{m_0+r}=E_r>0\qquad (r\ge 0).\) For each \(r\ge 0\), since \(e_{m_0+r}=f_{m_0+r}=E_r\), there exists an integer \(h_r\ge 0\) such that \(e_{m_0+r+1}=E_r+h_r.\) By (8), \(f_{m_0+r+1}=2E_r-h_r.\) Because equality persists, \(E_r+h_r=2E_r-h_r.\) Thus \(E_r=2h_r.\) Therefore \(E_{r+1}=e_{m_0+r+1}=E_r+h_r =E_r+\frac{E_r}{2} =\frac{3E_r}{2}.\) By induction, \(E_r=\left(\frac32\right)^rE_0=\frac{3^rE_0}{2^r}\) for every \(r\ge 0\). Since \(E_r\) is an integer and \(\gcd(3^r,2^r)=1\), this implies \(2^r\mid E_0\) for every \(r\ge 0\), impossible for a fixed positive integer \(E_0\). Hence equality \(e_t=f_t>0\) cannot persist forever.

Combining all cases, for each prime \(p\in\mathcal P\), the sequence \(v_p(X_m)=e_m\) is eventually constant.

Since \(\mathcal P\) is finite, there exists an index \(M\) such that for every \(p\in\mathcal P\), the value \(v_p(X_m)\) is constant for all \(m\ge M\). No prime outside \(\mathcal P\) divides any \(X_m\). Therefore the full prime factorization of \(X_m\) is the same for all \(m\ge M\), so \(X_m\) is constant for all \(m\ge M\).

But by (2), \(X_{m+1}=X_m+Y_m,\) and \(Y_m\) is a positive integer. Hence \(X_{m+1}>X_m\) for every \(m\ge 0\), contradicting the eventual constancy of \(X_m\).

Therefore no such positive integer \(N\) and infinite sequence of positive integers exist.

\paragraph{AI correctness audit.} Score: 7/7.
\begin{quote}
The proof is complete and correct. It derives the recurrence $X_{m+1}Y_{m+1}=X_mY_m^2$, confines all prime factors to a fixed finite set, and rigorously shows via $p$-adic valuation cases that every $v_p(X_m)$ eventually stabilizes. Finite prime support then makes $X_m$ eventually constant, contradicting $X_{m+1}=X_m+Y_m$ with $Y_m>0$.
\end{quote}

\paragraph{AI step audit.}
\begin{enumerate}
\item \textbf{Step 1: absent.} The submission constrains prime support by deriving divisibility such as $X_{m+1}\mid X_m^3$ and a finite prime set, but it never splits $X_m$ and $Y_m$ by their gcd into coprime parts. Thus it uses a materially different mechanism rather than recognizing the specified gcd/coprime-parts step.
\item \textbf{Step 2: present.} The submission explicitly isolates the persistent equality case $e_m=f_m=E>0$, assumes it persists, derives $E_{r+1}=3E_r/2$, and concludes $2^r\mid E_0$ for all $r$, which is impossible. This matches the valuation-recursion and fixed-exponent divisibility obstruction.
\item \textbf{Step 3: present.} The submission explicitly proves that once $f_m>e_m$, then $e_{m+1}=e_m$ and $f_{m+1}=2f_m>e_{m+1}$, so the strict valuation inequality propagates forward. It then applies eventual constancy uniformly over the finite prime set $P$ to force eventual constancy of $X_m$.
\end{enumerate}

\paragraph{Human audit.} Auditor B assigned 7/7 (``Sets up a recurrence and goes a slightly different way. Correct'') and marked step 1 absent and steps 2 and 3 present (``Step 1: does constrain prime support but differently Step 2: Idea present Step 3: Idea present'').

\end{document}